\documentclass{article}
\usepackage[preprint]{log_2022}
\usepackage[backref=page]{hyperref}
\usepackage{booktabs,multirow,amsfonts,graphicx,cleveref,dcolumn}
\usepackage{xcolor}

\usepackage[numbers,compress,sort]{natbib}
\usepackage{comments}
\Crefname{equation}{Eq.}{Eqs.}

\title[Learning Reduced-Order Models with GNNs]{Learning Reduced-Order Models for Cardiovascular Simulations with Graph Neural Networks}

\author[Pegolotti et al.]{%
Luca Pegolotti\\
\institute{Stanford University}\\
\email{lpego@stanford.edu}\And
Martin R. Pfaller\\
\institute{Stanford University}\\\And
Natalia L. Rubio\\
\institute{Stanford University}\\\And
Ke Ding\\
\institute{Intel Corporation}\\\And
Rita Brugarolas Brufau\\
\institute{Intel Corporation}\\\And
Eric Darve\\
\institute{Stanford University}\\\And
Alison L. Marsden\\
\institute{Stanford University}\\
}

\begin{document}

\maketitle

\begin{abstract}
Reduced-order models based on physics are a popular choice in cardiovascular modeling due to their efficiency, but they may experience reduced accuracy when working with anatomies that contain numerous junctions or pathological conditions.
%Physics-based reduced-order models are widely used in cardiovascular modeling for their efficiency. However, they sometimes suffer from loss of accuracy in anatomies featuring many junctions or pathological conditions. 
We develop one-dimensional reduced-order models that simulate blood flow dynamics using a graph neural network trained on three-dimensional hemodynamic simulation data. Given the initial condition of the system, the network iteratively predicts the pressure and flow rate at the vessel centerline nodes. Our numerical results demonstrate the accuracy and generalizability of our method in physiological geometries comprising a variety of anatomies and boundary conditions. 
Our findings demonstrate that our approach can achieve errors below 2\% and 3\% for pressure and flow rate, respectively, provided there is adequate training data. As a result, our method exhibits superior performance compared to physics-based one-dimensional models, while maintaining high efficiency at inference time.
%We show that, given adequate training data, our approach attains errors below 2\% and 3\% in pressure and flow rate, respectively, thus achieving superior performance compared to physics-based one-dimensional models without compromising efficiency in inference time.
\end{abstract}

% \begin{abstract}
% We develop one-dimensional reduced-order models to simulate blood flow dynamics using graph neural networks. First, we generate the training data by performing computationally expensive three-dimensional hemodynamic simulations. In a post-processing stage, we approximate the geometry of each artery with its centerlines and compute pressure and flow rate at each centerline node. We then train the network to approximate the increment in these quantities between consecutive timesteps. During the rollout phase of the method, the network acts as a solver, taking as input the previous system state prediction at each time step. We demonstrate the accuracy and potential for generalization of our implementation in physiological geometries comprising a variety of anatomies (aorta and aortofemoral models) and boundary conditions.
% \end{abstract}

% 

\section{Introduction}

In the last twenty years, Computational Fluid Dynamics (CFD) has become an essential tool in the study of the cardiovascular system \cite{bao2014usnctam,figueroa2017,schwarz2023beyond}. For example, CFD simulations have been used to noninvasively assess the severity of coronary artery aneurysms \cite{menon2022risk,gutierrez2019hemodynamic}, to propose novel surgical methods for congenital heart disease \cite{marsden2009evaluation}, and to optimize medical devices \cite{gundert2012optimization,sankaran2012patient}. Full 3D blood flow models, often solved with the finite element method, allow for patient-specific modeling and extraction of detailed quantities such as wall shear stress and velocity fields. 

However, the implementation of these simulations in clinical practice is still limited due, in part, to their high computational cost. Reduced-order models (ROMs) have been devised to overcome this issue, though their increased efficiency generally comes with a cost of lost accuracy. This paper aims to develop and validate a novel ROM following a data-driven approach based on graph neural networks (GNNs). 

The typical approach to deriving ROMs for cardiovascular simulations is physics-based. These formulations rely on simplifying assumptions that reduce the vessel geometry's complexity, and, consequently, the number of variables necessary to describe the quantities of interest. This physics-based class of ROMs includes popular zero-dimensional and one-dimensional models. 

In zero-dimensional formulations (often called lumped-parameter network models) the cardiovascular system is analogous to an electric circuit in which the pressure drop across portions of the arterial tree and blood flow rate play the role of electric potential difference and currents, respectively. These quantities of interest do not depend on any spatial variable. We refer to \cite{migliavacca2006multiscale,kung2013predictive, kim2010developing,kim2010patient} for examples of uses of such models in cardiovascular simulations, from simple Windkessel models \cite{westerhof2009arterial} to full circulatory networks. 

One-dimensional models are derived by integrating the three-dimensional Navier-Stokes equation over the vessel cross-section to reduce the equations to a single spatial variable \cite{hughes1973one}. Specifically, arterial trees are approximated as compositions of segments representing the centerline of the vessels, and pressure, flow rate, and vessel wall displacement are considered functions of the axial component only. Compared to lumped-parameter network models, one-dimensional models capture more of the physics due to their ability to account for wave propagation phenomena due to the interaction of flow with elastic vessel walls. 
%One-dimensional models can be coupled with three-dimensional ones to reach higher resolution in specific regions of the cardiovascular tree \cite{blanco2009potentialities,blanco2007unified,blanco20103d}. In \cite{Grande-Gutierrez:2022td}, for example, the authors follow this approach to study coronary hemodynamics. 
One-dimensional models have proven to be useful in numerous studies; see, for example, \cite{Grande-Gutierrez:2022td,reymond2013physiological,xiao2014systematic,hasan2021computationally,boileau2018estimating,blanco2018comparison,muller2021impact,moore2005one,grinberg2011modeling,reymond2009validation,reymond2012patient,boileau2015benchmark,bertaglia2020computational,blanco2020anatomical}. 

Zero- and one-dimensional formulations often lead to reasonably accurate results at a fraction of the computational cost of full three-dimensional simulations. However, these ROMs sometimes perform poorly because (i) they do not account correctly for pressure losses at vascular junctions and (ii) they rely on specific mathematical models (mostly based on empirical data or heuristics), particularly to describe pathological cases (e.g., stenoses or aneurysms). A recent study conducted on 72 cardiovascular models selected from a freely available database of cardiovascular models, the Vascular Model Repository (VMR),\footnote{\url{http://www.vascularmodel.com}} compared zero- and one-dimensional formulations against three-dimensional reference simulations \cite{pfaller2022}. The authors observed average relative errors of 2.1\% \& 1.8\% on the pressure (zero- and one-dimensional models, respectively) and 3.9\% \& 3.4\% on the flow rate (zero- and one-dimensional models, respectively).

Data-driven ROMs have the potential to overcome the issues mentioned above. %, but few attempts have been made to leverage the wealth of geometrical and simulation data generated during the past years by the cardiovascular modeling community. 
Typical data-driven reduced-order approaches include projection-based methods such as reduced-basis or proper-orthogonal decomposition techniques \cite{hesthaven2015certified,quarteroni2015reduced}. Although these methods can be adapted to specific geometries through interpolation strategies \cite{barrault2004empirical,negri2015efficient,santo2019hyper,manzoni2017efficient}, they are often not capable of representing the full range of geometrical variability characterizing patient-specific anatomic models. These problems can be partially overcome by employing projection-based methods in simple subdomains making up the vascular geometry of interest \cite{iapichino2016reduced, pegolotti2021model}.

Another option for generating data-driven models for physical simulations is deep learning.  While, in the context of cardiovascular modeling, deep learning has been used to accelerate other aspects of the simulation pipeline, for example, image-based segmentation \cite{maher2019accelerating,maher2020neural} and model generation \cite{kong2021whole,kong2020automating,kong2021deep}, it has not yet been used widely to model physical processes. Among the most popular architectures devoted to this scope are physics-informed neural networks (PINNs) \cite{raissi2017physicsI,raissi2017physicsII}, which are neural networks trained to mimic known physical laws. PINNs have been applied, for example, to ocean and climate modeling in \cite{lutjens2021pce,kashinath2021physics}, in the field of atomistic simulations in \cite{pun2019physically}, to approximate cardiac activation mapping in \cite{sahli2020physics} and blood flow dynamics in \cite{kissas2020machine} and \cite{arzani2021uncovering}. For an overview of fluid dynamics and machine learning efforts, we refer to \cite{brunton2020machine}. As with projection-based methods, these algorithms are typically not flexible with respect to changes in the domain geometry, which is a crucial drawback for their use in cardiovascular simulations on patient-specific applications. 

Recently, GNNs have been proposed as an alternative to classic fully connected and convolutional neural networks to address these difficulties. GNNs are used, for example, in \cite{sanchez2020learning} to learn the laws governing particle interactions in particle-based simulations and in \cite{pfaff2020learning} as solvers for mesh-based simulations called MeshGraphNets. More recently, GNNs have also been used to estimate steady blood flow in a synthetic dataset of three-dimensional arteries \cite{suk2023se}.

 In this paper, we consider an approach inspired by MeshGraphNets to derive a one-dimensional surrogate ROM for cardiovascular simulations. We apply the GNN to learn the laws governing bulk quantities, such as average pressure and flow rate at the centerline nodes, as opposed to using it to approximate the solution of the three-dimensional blood flow equations (as in the original paper). Owing to its capacity to resolve pressure and flow rate on cardiovascular models' centerlines, we consider our method a data-driven one-dimensional model. The network iteratively considers the state of the system, comprising pressure and flow rate at a particular timestep and other relevant features (such as cross-sectional area and parameters governing the boundary conditions), and computes approximations for the next values of pressure and flow rate. 
 
 We adapt the original MeshGraphNet implementation to make it suitable for blood flow simulations. In particular, we include graph edges to efficiently transfer boundary condition information to the interior graph nodes, and we include special parameters as node features to handle Windkessel-type or resistance-type boundary conditions, which are typically encountered in cardiovascular modeling. In our numerical results, we demonstrate the GNN on a diverse set of geometries. We assess its ability to generalize to multiple geometries and compare its performance with physics-based one-dimensional models.

In summary, the main contributions of this paper are threefold:
\begin{enumerate}
    \item We consider a modified version of MeshGraphNet capable of handling complex boundary conditions by adding special edges and patient-specific features to the graph. We show that these modifications lead to substantial improvements compared to MeshGraphNet in an ablation study performed in \Cref{subsec:comparison}. 
    \item We assess the ability of the GNN to generalize to different geometries, which, to our knowledge, has not been investigated in previous works. We reiterate that geometrical variability plays an essential role in patient-specific modeling (see  \Cref{subsec:convergence} and \Cref{subsec:comparison}).
    \item We demonstrate that our ROM outperforms physics-driven one-dimensional models in geometries characterized by many junctions or pathological conditions (see \Cref{subsec:comparison}).
\end{enumerate}

Our GNN implementation is freely available at \url{https://github.com/StanfordCBCL/gROM}.

\section{Notation}
\label{sec:notation}

Let us consider a set of $G$ patient-specific cardiovascular geometries $\Omega_1, \ldots, \Omega_G$. Given a geometry $\Omega_g$, we generate a directed graph consisting of a set of nodes $n^g_1, \ldots, n^g_{N^g}$ along the vessel centerline; see \Cref{fig:schematics_meshgraphnet} (left). We denote $e^g_{ij}$ as the directed edge connecting node $i$ to node $j$. We denote $T_\text{cc}^g$ as the period of one cardiac cycle of the patient and consider the discrete time sequence
\[
t^{0,g}, t^{1,g}, \ldots, t^{M,g},
\]
such that $t^{0,g} = 0$, $t^{M,g}= T_\text{cc}^g$, and $\Delta t^g = t^{1,g} - t^{0,g} = \ldots = t^{M,g} - t^{M-1,g}$. 

In this paper, we take a constant $\Delta t^g$ for all patients to simplify the learning process. However, we envision that the generalization to variable time step size is possible with minimal changes to the architecture (for instance, by including $\Delta t^g$ in the set of node features). 

We call the set of all node and edge features at time $t^{k,g}$ the state of the system $\Theta^{k,g}(\boldsymbol \mu)$, where $\boldsymbol \mu$ is a vector of system parameters---in this work, these are related to its boundary conditions.  We call a sequence of states $\Theta^{k,g}(\widetilde{\boldsymbol \mu})$ for a particular choice of system parameters $\widetilde{\boldsymbol \mu}$ a trajectory. We refer to \Cref{subsec:features} for a description of the features we consider in this work. For brevity, we will implicitly assume that all quantities must be patient-specific and omit the superscript $g$ in the following sections whenever possible.

\section{MeshGraphNets for cardiovascular simulations}
\label{sec:meshgraphnets}
\begin{figure}
  \centering
  \includegraphics[width = \textwidth]{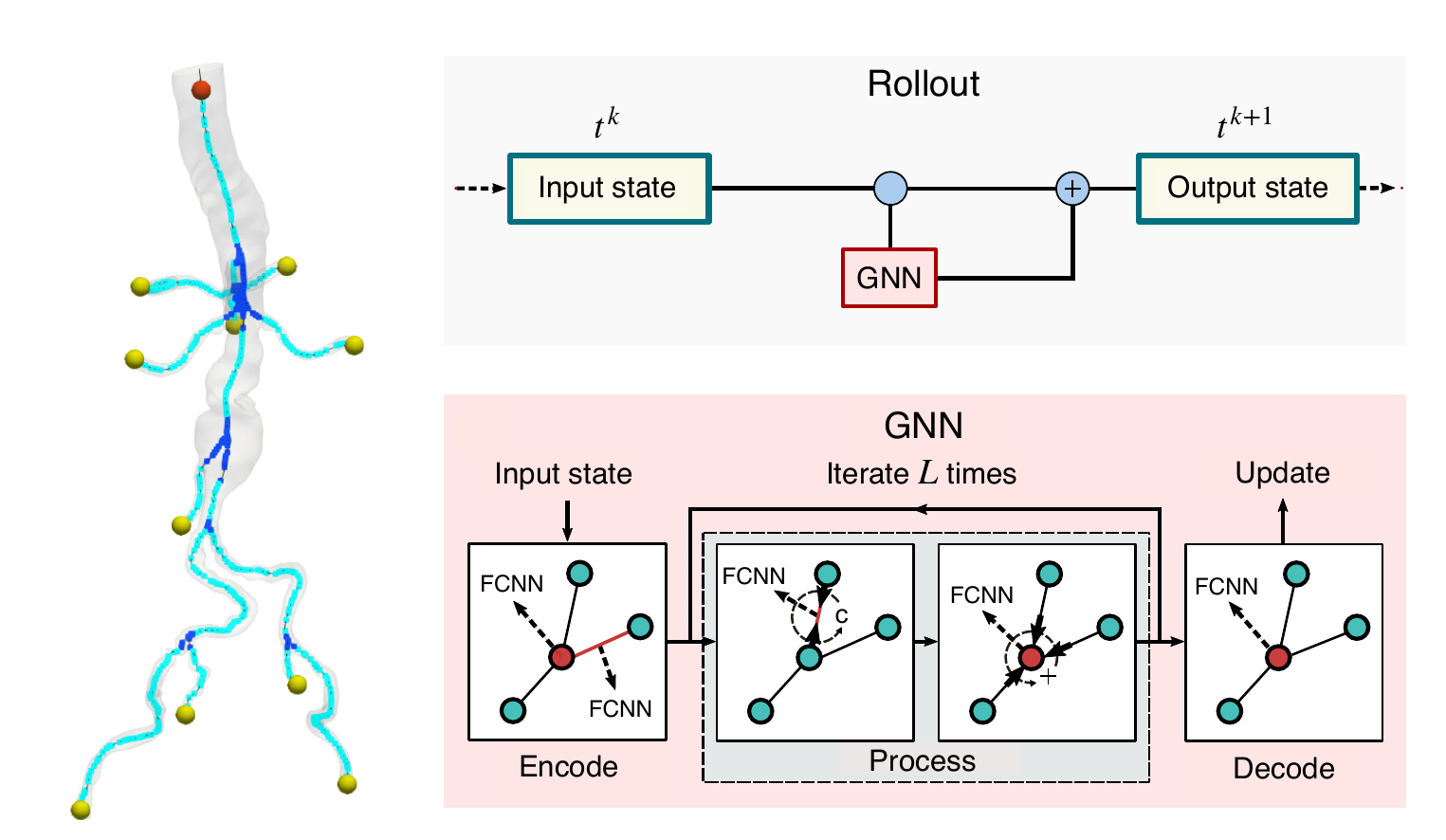}
  \caption{Schematics of MeshGraphNet. Left: graph of an aortofemoral model, with branch (light blue), junction (dark blue), inlet (red), and outlet (yellow) nodes. Top-right: rollout phase of the method. We provide the input state $\Theta^k$ to the GNN, which computes a prediction for the update of the state variables. The update is combined with the current state to compute $\Theta^{k+1}$. Bottom-right: steps in the GNN. The node and edge features are encoded using fully-connected neural networks (FCNNs), processed $L$ times using aggregation operations, and decoded into the output space.}
  \label{fig:schematics_meshgraphnet}
\end{figure}

Our GNN acts as a data-driven one-dimensional ROM; for a description of classical physics-based one-dimensional models, we refer to \Cref{sec:oned}. During the rollout phase shown in \Cref{fig:schematics_meshgraphnet} (top-right), the network takes as an input $\Theta^k$ and computes an update that allows us to advance the state of the system from $\Theta^k$ to $\Theta^{k+1}$. We apply the GNN iteratively. In each time step $t^k$ with $k > 0$ we provide it with the previously estimated system state. At $t^0$, we feed the network a prescribed initial condition. The action of the GNN is described in \Cref{subsec:forward_gnn}.

\subsection{Graph features}
\label{subsec:features}
We select the node and edge features based on our knowledge of the problem. For example, it is well known that, under a Poiseulle condition, a linear relation exists between flow rate $Q$ and pressure drop $\Delta P$ across an approximately cylindrical vessel. Specifically,
\[
    \Delta P = R Q = \dfrac{8 \mu L}{\pi r^4} Q.
\]
The proportionality constant $R$ is called resistance and depends on the viscosity of blood $\mu$, the length of the vessel $L$, and its radius $r  $. In this work, we assume that the viscosity and density of the blood are equal to $\mu = 0.04\text{ g}\,\text{cm}^{-1}\,\text{s}^{-1}$ and $\rho = 1.06\text{ gr}\,\text{cm}^{-3}$ for all patients, respectively, and we do not include viscosity and density as graph features. Based on these notions of fluid dynamics, we include the cross-sectional area in the node features.

\paragraph{Node features} We consider the cross-sectional average pressure $p^k_i \in \mathbb{R}^{+}$ and the flow rate $q^k_i \in \mathbb{R}$ in every centerline node $n_i$ as descriptors of the state of the system $\Theta^k$ at time $t^k$. These are computed on the section obtained as the intersection between the plane orthogonal to the centerline and the thee-dimensional model of the vessel. Similarly, the area of the vessel lumen $A_i \in \mathbb{R}$ is the area of the section passing through node $i$ and is considered a node feature for the above reasons. We introduce a one-hot vector $\boldsymbol{\alpha}_i \in \mathbb{R}^{4}$ to encode different node types. These are branch nodes, junction nodes, model inlet (one per geometry), and model outlets, as shown in \Cref{fig:schematics_meshgraphnet} (left). The distinction between branch and junction nodes is necessary because the area of sections within junctions varies discontinuously across the centerline (owing to nearby slices sometimes cutting through a different number of branches). This leads to different blood dynamics in branches and junctions when looking at quantities averaged over such sections. We use the same automated algorithm for junction detection as in \cite{pfaller2022}. Additional node features are the tangent to the centerline evaluated at node $n_i$, which we denote $\boldsymbol{\phi}_i \in \mathbb{R}^3$, the period of the cardiac cycle $T_\text{cc} \in \mathbb{R}^+$, minimum and maximum pressure over the whole cardiac cycle ($p_\text{min}\in \mathbb{R}^+$ and $p_\text{max}\in \mathbb{R}^+$), three parameters associated with the boundary conditions ($R_{i,p}\in \mathbb{R}^+$, $C_i\in \mathbb{R}^+$, and $R_{i,d}\in \mathbb{R}^+$; see \Cref{subsec:bcs} for further information), and a boolean loading variable $l^k \in \{0,1\}$ associated with model initialization (see \Cref{subsec:initialization}). We remark that, although in this work $p_\text{min}$ and $p_\text{max}$ are set based on simulation data, these values are typically known for the patient at hand---for example, they are identified as diastolic and systolic pressure in aorta models---and are often used in physics-based cardiovascular simulations to tune the boundary conditions discussed in \Cref{subsec:bcs}. In summary, we associate with each node $n_i$ the vector of features 
\begin{equation}
    \mathbf{v}^k_i = [p_i^k, q_i^k, A_i, \boldsymbol{\alpha}_i^\text{T}, \boldsymbol{\phi}_i^\text{T}, T_\text{cc}, p_\text{min}, p_\text{max}, R_{i,p}, C_i, R_{i,d}, l^k]^\text{T} \in \mathbb{R}^{17}.
    \label{eq:node_features}
\end{equation}

Node features and their definitions are summarized in \Cref{table:node_features}.

\begin{table}[t]
\centering
\begin{tabular} {cl}
\toprule
Node feature & Definition \\
\midrule
$p_i^k$ & Pressure at time $t^k$\\
$q_i^k$ & Flow rate at time $q^k$\\
$A_i$ & Cross-sectional area\\ 
$\boldsymbol{\alpha}_i$ & Nodal type \\
$\boldsymbol{\phi}_i$ & Centerline tangent\\ 
$T_\text{cc}$ & Cardiac cycle duration\\
$p_\text{min}$ & Minimum pressure (across cardiac cycle)\\ 
$p_\text{max}$ & Maximum pressure (across cardiac cycle)\\
$R_{i,p}$ & $R_p$ Parameter in RCR boundary conditions (see \Cref{fig:RCR}) \\
$C_i$ & $C$ Parameter in RCR boundary conditions (see \Cref{fig:RCR}) \\
$R_{i,d}$ & $R_d$ Parameter in RCR boundary conditions (see \Cref{fig:RCR}) \\
$l^k$ & Boolean load variable at time $t^k$ \\\bottomrule
\end{tabular}
\caption{Node features and their definitions.}
\label{table:node_features}
\end{table}

\paragraph{Edge features} We define $\mathbf{d}_{ij} = \mathbf{x}_j - \mathbf{x}_i \in \mathbb{R}^3$ as the difference between the position of nodes $n_j$ and $n_i$, and $z_{ij}$ the length of the shortest path between the two nodes. In addition to \textit{physical} edges, in this work we introduce edges connecting boundary nodes to interior ones (see \Cref{subsec:bcs}). We, therefore, introduce the one-hot vector $\boldsymbol{\beta}_{ij} \in \mathbb{R}^{4}$ to encode the edge type. We define the following edge types: edges connecting branch nodes to branch nodes, edges connecting junction nodes to junction nodes, and edges connecting the model inlet or outlets to interior nodes. For simplicity, edges connecting branch nodes to junction nodes are assigned the same type as those within branches. We note that edges within branches and junctions are those defining the centerline of the anatomical model and that, when computing the shortest path $z_{ij}$, we only consider these edges. Moreover, we highlight that although the introduction of boundary edges changes the topology of the graph, information regarding the original topology is maintained through the different edge types. The edge features associated with $e_{ij}$ are
\begin{equation}
 \mathbf{w}_{ij} = [\mathbf{d}_{ij}^\text{T} / \Vert \mathbf{d}_{ij} \Vert_2, z_{ij}, \boldsymbol{\beta}_{ij}^\text{T}]^\text{T} \in \mathbb{R}^8.
  \label{eq:edge_features}
\end{equation}

\begin{table}[t]
\centering
\begin{tabular} {cl}
\toprule
Edge feature & Definition \\
\midrule
$\mathbf{d}_{ij}^\text{T} / \Vert \mathbf{d}_{ij}$ & Normalized vector distance between node $i$ and $j$\\
$z_{ij}$ & Shortest path length between node $i$ and $j$\\
$\boldsymbol{\beta}_{ij}$ & Edge type\\ 
\bottomrule
\end{tabular}
\caption{Edge features and their definitions.}
\label{table:edge_features}
\end{table}

Edge features and their definitions are summarized in \Cref{table:edge_features}.

Note that all features (except for the unitary vectors $\mathbf{t}_i$ and $\mathbf{d}_{ij}^\text{T} / \Vert \mathbf{d}_{ij} \Vert_2$) are normalized to follow a standard Gaussian distribution $\mathcal{N}(0,1)$ using statistics computed on the dataset.

\subsection{Extension of MeshGraphNet to support cardiovascular boundary conditions}
\label{subsec:bcs}

\begin{figure}
  \centering
  \includegraphics[scale = 0.4]{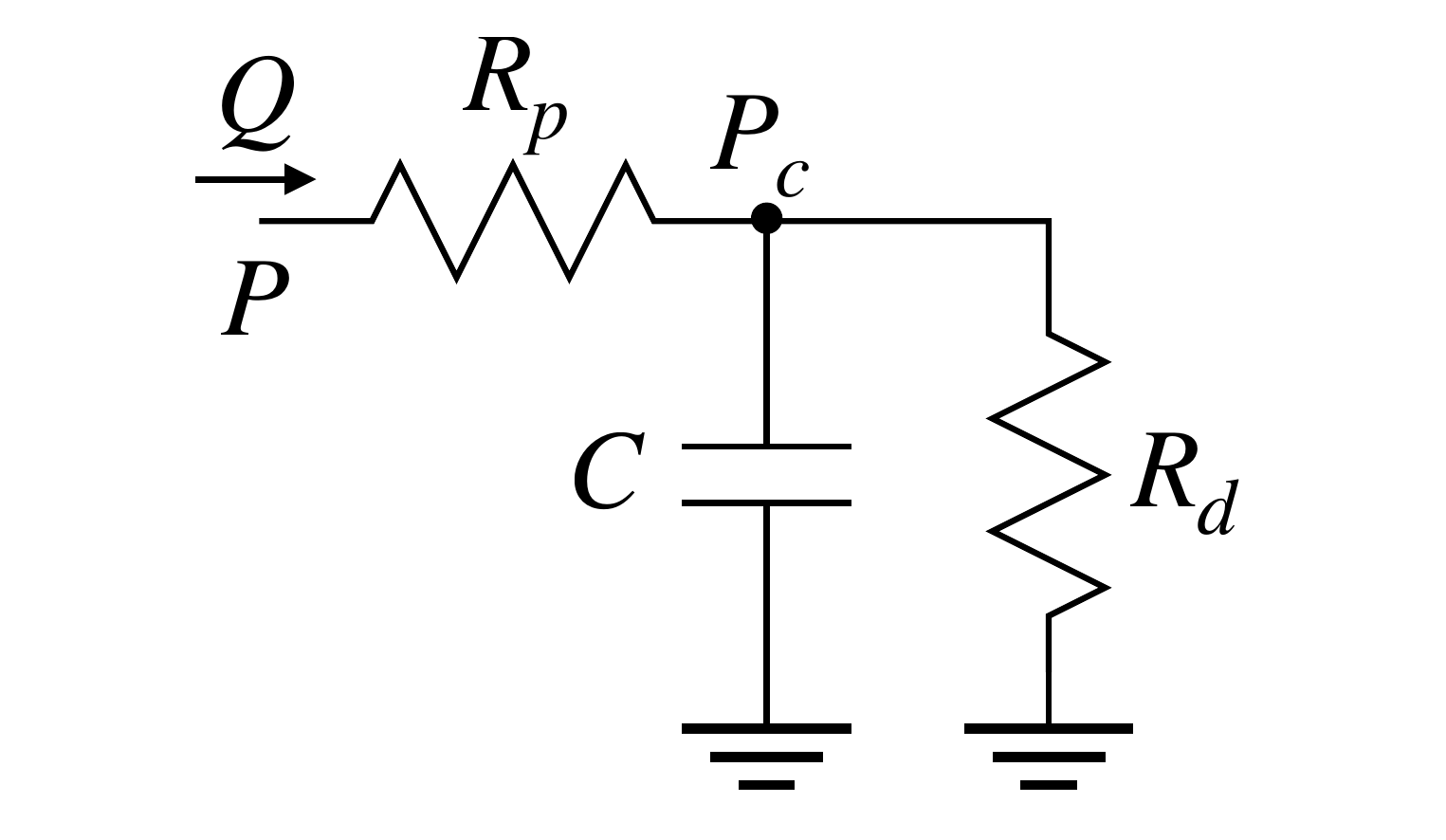}
  \includegraphics[scale = 0.4]{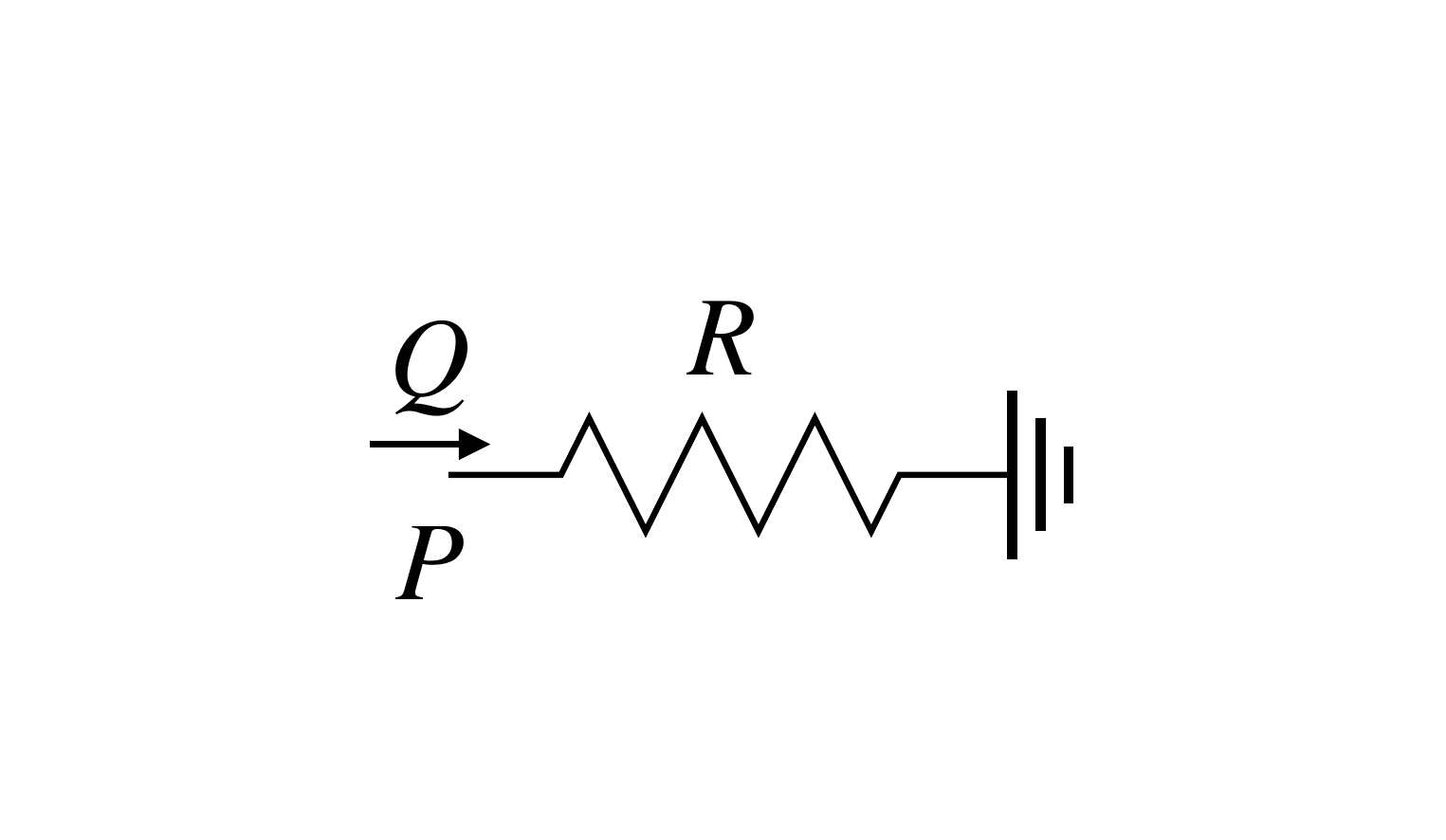}
  \caption{Circuit schematics of three-element Windkessel (RCR) and resistance boundary conditions (left and right, respectively). In physics-based models these boundary conditions are modeled using ordinary differential equations such as \Cref{eq:rcr}. Here, we train the GNN to characterize these boundary conditions by including $R_p, C, R_d$ or $R$ into the graph features.}
  \label{fig:RCR}
\end{figure}

The hemodynamics in the vasculature are determined by the boundary conditions at the inlet and outlets. In cardiovascular simulations, we represent the downstream vasculature as three-element Windkessel-type (RCR) or resistance-type boundary conditions, as shown in \Cref{fig:RCR}. 

In RCR boundary conditions, the pressure in the capacitor $P_c(t, Q)$ is determined by the relationship
\begin{equation}
\dfrac{\text{d} P_c}{\text{d} t} = \dfrac{1}{C}\left (Q(t) - \dfrac{P_c}{R_d} \right ).
\label{eq:rcr}
\end{equation}
Then, the pressure at the outlet ($P$ in \Cref{fig:RCR}) is found as $P(t, Q) = P_c(t,Q) - R_p Q(t)$. 

In resistance boundary conditions, the relationship between pressure and flow rate at the outlet is simply $P(t) = R Q(t)$. We remark that this is a special case of RCR boundary conditions where $C$ and $R_d$ tend to zero. 

In both RCR and resistance boundary conditions, we assume that the distal pressure is zero for simplicity.

To integrate these boundary condition types into the MeshGraphNet framework, we provide the values of $R_p$, $C$, and $R_d$ as node features in the outlet nodes. We set those features to zero for all remaining nodes in the graph. When using resistance-only boundary conditions, we also set $C = 0$ and $R_d = 0$.
 During training and in every iteration of the rollout phase, we solely prescribe the flow rate value at the next time step at the inlet. 
 
 Due to the nature of the centerlines we consider in this paper, most centerline nodes are connected to only two neighbors, which hinders the information transfer between the boundary nodes and the interior ones. For this reason, we add artificial edges connecting each interior node to the closest node on the boundary. Specifically, every interior node $n_i$ is connected to the boundary node $n_j$ for which the shortest path length between $n_i$ and $n_j$, i.e., $z_{ij}$, is minimized. Edges connecting inlets and outlets to interior nodes are associated with different types of edges (see \textit{Edge features} in \Cref{subsec:features}) and are bidirectional as all the other types of edges that we consider in this paper.  

\subsection{Initial conditions}
\label{subsec:initialization}

As we shall see in \Cref{sec:dataset}, we train our GNN on time-dependent simulation data over one cardiac cycle. However, the initial condition of the system at the start of the cardiac cycle is in general not a piece of information that can be easily estimated from clinical data. For this reason, when generating the trajectories in the dataset, we start from a constant solution in which $p^0 = p_\text{min}$ and $q^0 = 0$ in all graph nodes and then linearly interpolate between this initial state and the actual initial condition at the start of the cardiac cycle over $T_l = 0.1$ s. We then include a node feature,  denoted $l^k$, taking a unitary value between $t = -0.1$ sec and $t = 0$ sec, and zero for $t > 0$, to differentiate between the loading stage and the actual simulation of the cardiac cycle.

\subsection{Forward step of the Graph Neural Network}
\label{subsec:forward_gnn}
In this section, we denote by $\mathbf{f}_{*}$ a fully-connected neural network (FCNN) \cite{goodfellow2016deep} with $n_h$ hidden layers. The number of neurons in the hidden layers is constant and equal to $n_s$. We consider the LeakyReLU activation function for every layer except the output layer, and we apply a layer normalization on the output layer unless explicitly specified. When an FCNN accepts multiple arguments, these should be concatenated into a single tensor.

MeshGraphNet is based on the three main stages shown in \Cref{fig:schematics_meshgraphnet} (bottom-right):
\begin{enumerate}
\item \textit{Encode}: we transform the node and edge features into latent features using FCNNs. In particular, given node feature $\mathbf{v}_i^k$, we compute its latent representation as $\mathbf{v}_i^{k,(0)} = \mathbf{f}_{\text{en}}(\mathbf{v}_i^k) \in \mathbb{R}^{n_l}$, where $\mathbf{f}_{\text{en}}$ is an FCNN mapping the space of node features into $\mathbb{R}^{n_l}$. Similarly, we compute $\mathbf{w}_{ij}^{(0)} = \mathbf{f}_{\text{ee}}(\mathbf{w}_{ij}) \in \mathbb{R}^{n_l}$ for all edges $e_{ij}$. Although the size of the latent space $n_l$ does not need to coincide for edges and nodes, we take it to be the same to facilitate the optimization of our architecture, discussed in \Cref{subsec:hpo}. % We find the latent representations of boundary edges as $\tilde{\mathbf{w}}^{n}_{\text{in},j} = \mathbf{f}_\text{in}(\mathbf{w}_{\text{in},j},\mathbf{v}^k_{\text{in}}) \in \mathbb{R}^{n_{bc}}$ and $\tilde{\mathbf{w}}^{n}_{\text{out},j} = \mathbf{f}_\text{out}(\mathbf{w}_{\text{out},j},\mathbf{v}^n_{\text{out},i}) \in \mathbb{R}^{n_{bc}}$, where $i^\text{th}$ is the closest outlet in graph distance to node $n_j$.
\item \textit{Process}: the process stage is performed $L$ times and is further divided into two phases. In the first phase, we compute new edge features as
\[
\mathbf w_{ij}^{(l)} = \mathbf{f}_{\text{pe}}^{(l)}(\mathbf{w}_{ij}^{(l-1)}, \mathbf{v}_{i}^{k,(l-1)},  \mathbf{v}_{j}^{k,(l-1)})\in \mathbb{R}^{n_l},
\]
where $l \geq 1$ is the iteration number. Then, we compute new node features using aggregation functions as
\[
\mathbf v_{j}^{k,(l)} = \mathbf{f}_{\text{pn}}^{(l)}(\mathbf{v}_{j}^{k,(l-1)},  \sum_{i: \exists e_{ij}} \mathbf{w}_{ij}^{(l)}, \mathbf{w}_{\text{in},j}, \mathbf{w}_{\text{out},j})\in \mathbb{R}^{n_l}.
\]
All the networks considered in this stage feature a residual connection.
\item \textit{Decode}: in the last stage, node features are transformed from the latent space to the desired output space using an FCNN. The desired output is a vector that contains the update of pressure and flow rate $[\delta p_i^k, \delta q_i^k] = \mathbf{f}_\text{dn}(\mathbf v_{i}^{k,(L)}) \in \mathbb{R}^2$. Following \cite{pfaff2020learning}, we do not use layer normalization on the output layer of $\mathbf{f}_\text{dn}$.
\end{enumerate}
At the end of the forward pass of the GNN, we update the pressure and flow rate nodal values $p_i^k$ and $q_i^k$ as $p_i^{k+1} = p_i^k + \delta p_i^k$ and $q_i^{k+1} = q_i^k + \delta q_i^k$.

We introduce $\delta \mathbf{v}_i^k = [\delta p_i^k, \delta q_i^k, 0,\ldots,0] \in \mathbb{R}^{17}$ and the function $\Psi_m$, which denotes the result of applying the GNN $m$ consecutive times (rollout phase). Specifically,
\begin{align*}
\Psi_1(\Theta^k) & = \bigcup_{i = 1}^N \{ \mathbf{v}_i^k + \delta \mathbf{v}_i^k \}  \; \cup \!\! \bigcup_{i,j: \exists e_{ij}} \{\mathbf{w}_{ij}\},
\intertext{and}
\Psi_m(\Theta^k) & = (\underbrace{\Psi_1 \circ \cdots \circ \Psi_1}_{m})(\Theta^k).
\end{align*}
We denote the approximation of $p$ and $q$ at node $i$, after $m$ applications of the GNN, as $\Psi_m(\Theta^k)|_{p,i}$ and $\Psi_m(\Theta^k)|_{q,i}$, respectively. 

% We note that, as we discuss in more detail in \Cref{sec:dataset}, the blood dynamics in nodes belonging to junctions is more complex than that in nodes belonging to branches. Therefore, we propose to consider a heterogeneous graph in which node indices are partitioned into branch nodes indices $\mathcal N_b$ and junction nodes indices $\mathcal N_j$. FCNNs operating on branch or junction nodes (e.g., $\mathbf{f}_{\text{en}}$) have different weights. Similarly, FCNNs operating on edges feature different weights whether they are used on branch-branch, junction-junction, branch-junction and junction-branch edges.

\subsection{Training}
\label{subsec:training}
% \begin{figure}
%   \centering
%   \includegraphics{images/junctions_scheme.pdf}
%   \caption{\textbf{Junction handling.} (a) Aorta and corresponding centerlines. We color nodes with respect to whether they belong to branches (light blue) or junctions. (b) Strategy to compute mass loss at junctions. We first compute the average flow rate in branches using artificial nodes (one per branch, pink nodes in figure) and then compute the net flow rate loss using artificial nodes in junctions (one per junction, blue node in figure).}
%   \label{fig:schematics_junctions}
% \end{figure}
We now introduce the loss function $\mathcal L$ minimized during training of the GNN. Let us define the training set $\{\Omega_g\}_{g \in \mathcal T}$, where $\mathcal T$ is a subset of $[1,\ldots,G]$. We train the network on small strides of $s$ consecutive time steps so that it approximates the exact values of the nodal pressure and flow rate $\hat{p}^{k,g}_i$ and $\hat{q}^{k,g}_i$. Here, we consider $s = 5$. Considering multiple steps has been proven to be beneficial in \cite{wu2022learning}. In particular, we define the strided MSE
\[
    \text{sMSE}^{k,g,s} = \dfrac{1}{N^g} \sum_{l = 1}^{s} \sum_{i = 1}^{N^g}a_l b_i \left [ ( \hat{p}^{k+l,g}_i - \Psi_l(\Theta^{k,g})|_{p,i})^2 + (\hat{q}^{k+l,g}_i - \Psi_l(\Theta^{k,g})|_{q,i})^2 \right ],
\]
where $a_l$ = 1 if $l = 1$ and $a_l = 0.5$ otherwise, and $b_i = 100$ for boundary nodes and $b_i = 1$ otherwise. 

Then, the loss function reads
\begin{equation}
\mathcal L = \text{MSE} = \sum_{g \in \mathcal{T}} \dfrac{1}{|\mathcal{T}|(M^g-s)} \sum_{k = 0}^{M^g-s} \text{sMSE}^{k,g,s}.
\label{eq:loss}
\end{equation}

Our goal is to make the GNN robust for rollouts of many time steps. In order to achieve this, it is imperative to add random noise during training to simulate the effect of the error caused by the network in the rollout phase, as described in \cite{pfaff2020learning, sanchez2020learning}. During training, we perturb the pressure and flow rate in the state $\Theta^{k,g}$ as $\tilde{p}_i^g = \hat{p}_i^g + \varepsilon_p$ and $\tilde{q}_i^g = \hat{q}_i^g + \varepsilon_q$, where $\varepsilon_p \sim N (0,\sigma^2)$, $\varepsilon_q \sim N(0,\sigma^2)$, and $\sigma$ is a hyperparameter controlling the noise standard deviation. The loss function~\eqref{eq:loss} is optimized using stochastic gradient descent and the Adam optimizer. For details on the GNN and training hyperparameters, we refer to \Cref{subsec:hpo}.

We consider each entire trajectory as a datapoint. Due to the limited size of the datasets that we obtain in this way (as opposed to, for example, considering every possible set of $s$ consecutive timestep as separate datapoints), we use $k$-fold cross-validation~\cite{hastie2009elements}, that is, we train $k$ networks on different train-test splits using the same set of hyperparameters and report their average performance. The train-test splits are made such that $1-1/k$ and $1/k$ of the trajectories are assigned to the train and test set, respectively, and such that each trajectory appears in the test set exactly once. Consider as an example the dataset employed in \Cref{subsec:convergence}, which is composed of 160 trajectories. During 10-fold cross-validation, each of the 10 networks is trained on 144 trajectories and tested on 16, and we report the average performance achieved on train and test.

\subsection{Hyperparameter optimization}
\label{subsec:hpo}
Hyperparameter optimization is essential for achieving state-of-the-art performance using machine learning methods. In this work, we employ the optimization platform SigOpt.\footnote{\url{https://sigopt.com}} As our objective function, we employed the average rollout error on pressure and flow rate obtained on the test set after training. More specifically, given the anatomy of patient $g$, the errors for pressure and flow rate are computed as
\begin{align}
    \text{e}_p^g &= \dfrac{\sum_{i \in \mathcal{B}^g} \sum_{k = 1}^{M^g} (\hat{p}^{k,g}_i - \Psi_k(\Theta^{0,g})|_{p,i})^2}{\sum_{i \in \mathcal{B}^g} \sum_{k = 1}^{M^g} (\hat{p}^{k,g}_i )^2}\label{eq:err_p}, \\
    \text{e}_q^g &= \dfrac{\sum_{i \in \mathcal{B}^g} \sum_{k = 1}^{M^g} (\hat{q}^{k,g}_i - \Psi_k(\Theta^{0,g})|_{q,i})^2}{\sum_{i \in \mathcal{B}^g} \sum_{k = 1}^{M^g} (\hat{q}^{k,g}_i )^2}\label{eq:err_q},
\end{align}
where $\hat{p}^{k,g}_i$ and $\hat{q}^{k,g}_i$ are the exact pressure and flow rate values at time $t^k$ and at node $n_i$ for patient $g$. Note that, when computing these errors, we only consider the indices of nodes located in branches, that is, $\mathcal{B}^g$.

Our optimization yielded the following choices of hyperparameters:

\begin{enumerate}

\item \textit{Network architecture.} Each fully-connected network in the GNN consists of $n_h = 2$ hidden layers and $n_s = 64$ neurons per hidden layer. The optimal number of processing iterations in the forward step of the GNN is $L = 5$, and we consider $n_l = 16$ as the size for the latent space.

\item \textit{Training hyperparameters.} We consider a starting learning rate of $10^{-3}$. This is decreased using a cosine annealing function; the final value of learning rate is $10^{-6}$. We use batches of 100 graphs. During training, we inject normally distributed noise $N(0,\sigma^2)$ with $\sigma = 5 \cdot 10^{-2}.$ Unless explicitly stated, we train the GNNs for 100 epochs.
\end{enumerate}

% \cmt{could we provide more info here? usually hyperparameter tuning leads to some plots that summarize the performance, etc, the sensitivity of the parameters, etc.}

\section{Dataset}
\label{sec:dataset}

\begin{figure}
\centering
\includegraphics[width = 1\textwidth]{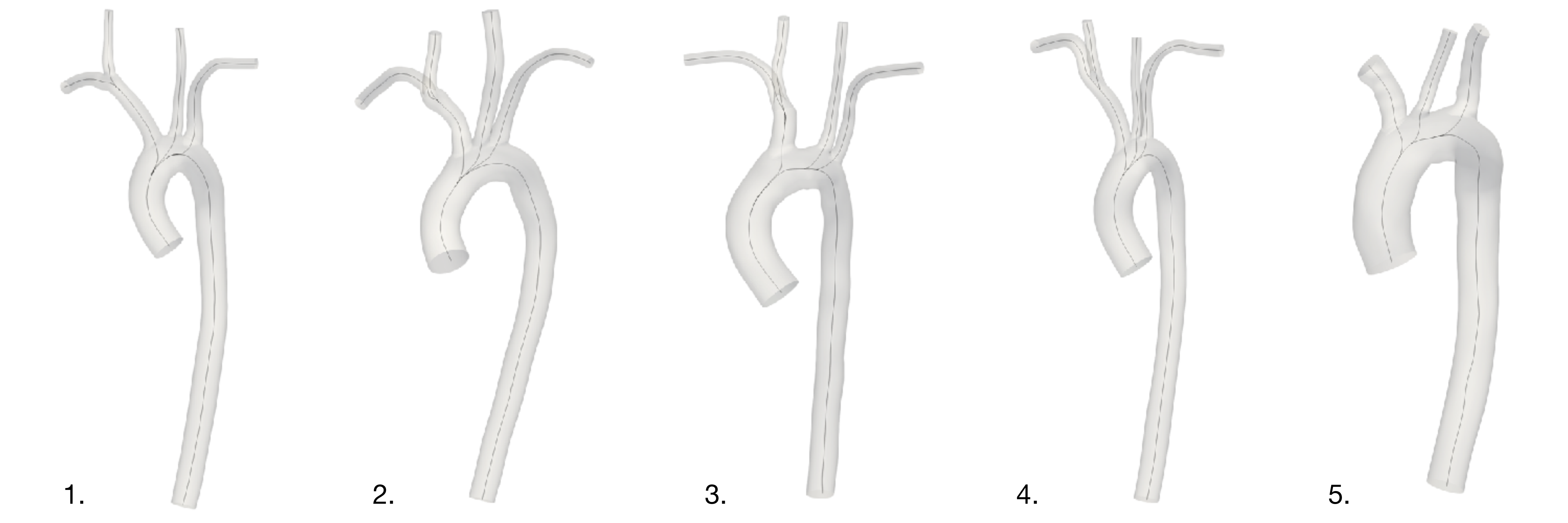}\\
\includegraphics[width = 1\textwidth]{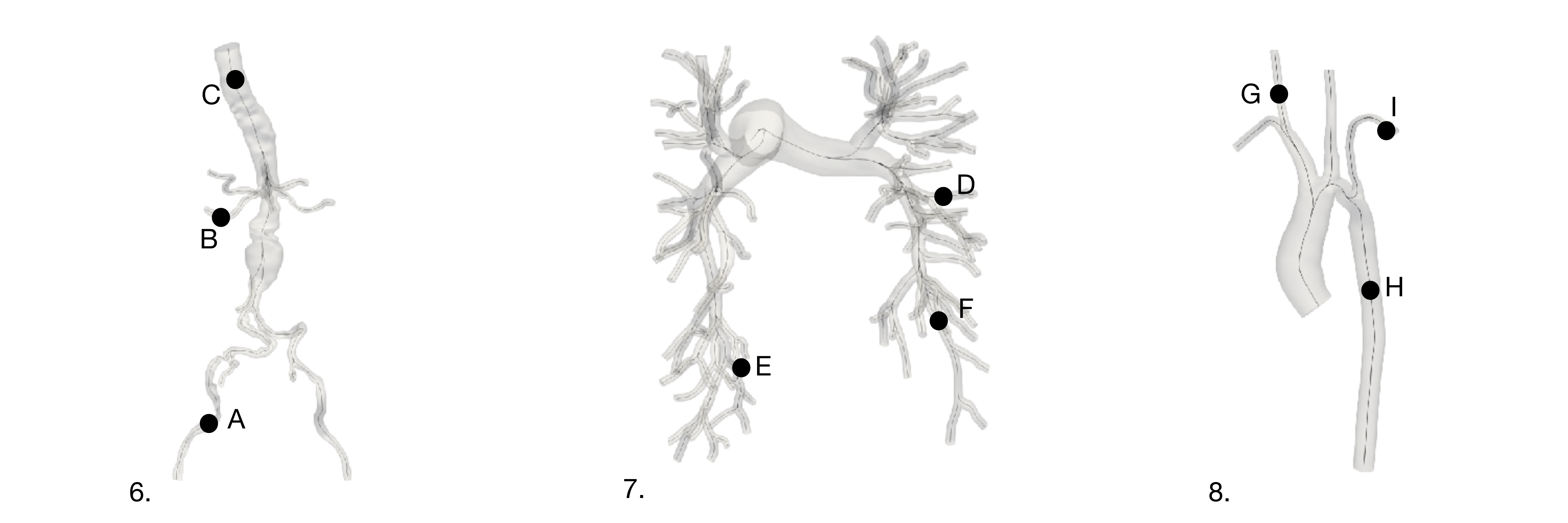}
\caption{Cardiovascular models from the Vascular Model Repository. Models 1-5: healthy aorta models. Model 6: aortofemoral model affected by an aneurysm. Model 7: healthy pulmonary model. Model 8: aorta model affected by coarctation. The locations marked in the bottom geometries are used in the results presented in \Cref{subsec:comparison}}
\label{fig:dataset}
\end{figure}

We consider eight healthy and diseased patient-specific models selected from the VMR and shown in \Cref{fig:dataset}. Models 1-5 are healthy aorta models, Model 6 is an aortofemoral model featuring an aneurysm, Model 7 is a healthy pulmonary model, and Model 8 is an aorta affected by coarctation. We selected these models as, based on previous studies such as \cite{pfaller2022}, they present features---e.g, many junctions (Model 7) or stenoses (Model 8)---that are typically challenging for traditional physics-based models.

The data-generation pipeline consists of three main steps: three-dimensional simulation using SimVascular \cite{updegrove2017simvascular,lan2018re}, transformation to the one-dimensional representation, and graph generation.

\paragraph{Three-dimensional simulation.} In addition to geometrical information, the VMR also contains simulation data and boundary conditions information tuned to match clinical measurements. The models include a prescribed flow rate profile over one cardiac cycle at the inlet and RCR (Models 1-6 and Model 8) or resistance boundary conditions (Model 7) at the outlets. We performed 32 (Models 1-5) and 50 (Models 6-8) three-dimensional simulations for each of these geometries under random perturbations of the original boundary conditions. In particular, we multiplied the inlet flow rate and each of the parameters governing the outlet boundary conditions by independent factors uniformly distributed in the range $[0.8,1.2]$. Using the notation introduced in \Cref{sec:notation}, each model in the VMR was associated with a particular parameter set $\boldsymbol \mu = [\mu^\text{in},\mu^\text{out}_1,\ldots,\mu^\text{out}_{N_\text{out}}]$ governing its boundary conditions, and we ran simulations for multiple perturbed parameter sets each of the form $\widetilde{\boldsymbol \mu} = [c^\text{in}\mu^\text{in} ,c^\text{out}_1\mu^\text{out}_1 ,\ldots,c^\text{out}_{N_\text{out}}\mu^\text{out}_{N_\text{out}}]$, where $c^\text{in},c^\text{out}_1,\ldots,c^\text{out}_{N_\text{out}} \sim U(0.8,1.8)$. We performed the three-dimensional finite-element simulations of the unsteady Navier--Stokes of two cardiac cycles on 128 dual-socket AMD(R) EPYC 7742 cores of the San Diego Super Computing Center (SDSC) Expanse cluster.

 In \Cref{table:3D-runtime} we report the average simulation run time for each of the eight considered models.

\begin{table}[t]
\centering
\begin{tabular} {c c c c c c c c c}
\toprule
 & 1 & 2 & 3 & 4 & 5 & 6 & 7 & 8 \\
\midrule
Run time [h] & 3.8 & 2.7 & 1.7 & 2.9 & 3.7 & 19.4 & 5.8 & 5.9\\
Number of simulations & 32 & 32 & 32 & 32 & 32 & 50 & 50 & 50\\
\bottomrule
\end{tabular}
\caption{Average run time in hours of three-dimensional simulations and total number of simulations for each of the considered cardiovascular models.}
\label{table:3D-runtime}
\end{table}

\begin{figure}
  \centering
  \includegraphics[trim={0 1cm 0 0.5cm},clip, width = \textwidth]{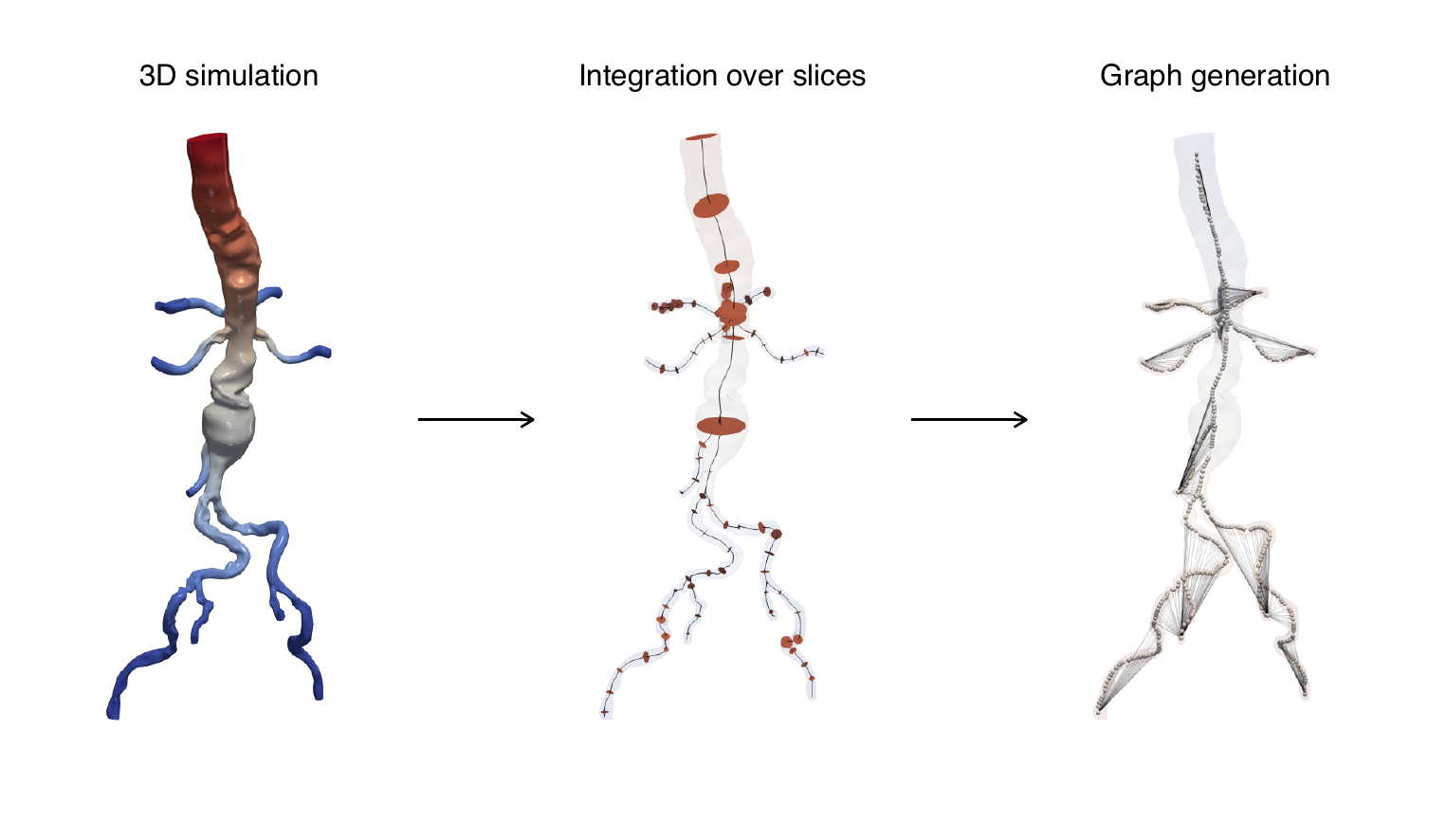}
  \caption{Pipeline for generating input data to MeshGraphNet. We first run three-dimensional simulations using SimVascular, then restrict the results to the centerlines by integrating pressure and velocity over slices, and finally generate graphs and relative node and edge features starting from the centerline nodes and connectivity. The figure on the right shows the boundary edges discussed in \Cref{subsec:bcs}.}
  \label{fig:pipeline}
\end{figure}

\paragraph{Post-processing of simulation results.} We restricted the three-dimensional results to the model centerlines. We achieved this by considering the orthogonal sections of the vascular geometry at each centerline node and integrating the pressure and normal component of the velocity over the cross section, thus computing average pressure and flow rate. We also associated with each centerline node the area of the corresponding section. For clarity, \Cref{fig:pipeline} (middle) shows a subset of the sections where we integrated the pressure and velocity field.

\paragraph{Graph generation. } Our GNN implementation is based on PyTorch and the Deep Graph Library (DGL)\footnote{\url{https://www.dgl.ai}}. In this last step of the pipeline, we generated nodes, edges, and relative features, in a format compatible with DGL. As discussed in \Cref{subsec:bcs} and shown in \Cref{fig:pipeline} (right), at this stage we added edges connecting boundary nodes to interior ones. The time-dependent pressure and flow rate fields are here resampled at each graph node using cubic splines at a constant $\Delta t$. We also performed data augmentation and increase the number of graphs 4-fold by starting each trajectory at a different offset. \Cref{table:statistics} shows statistics computed over the graphs considered in this paper. Each row is indexed based on the ID of the corresponding cardiovascular model (see \Cref{fig:dataset} for reference). 

These three steps of the pipeline are summarized in \Cref{fig:pipeline}.

\begin{table}[t]
\centering
\begin{tabular} {c c c c c c c}
\toprule
ID & $p$ [mmHg] & $q$ [cm$^2$/s] & $A$ [cm$^2$] & $h$ [mm] & nodes & edges \\
\midrule
1 & 84 [46,148] & 35 [-43,449] & 1.23 [0.2,3.3] & 2.06 [1.1,4.2] & 288 & 1138 \\
2 & 72 [42,209] & 15 [-20,376] & 0.55 [0.1,1.4] & 1.36 [0.8,3.0] & 292 & 1154 \\
3 & 74 [33,120] & 17 [-10,235] & 0.78 [0.1,2.6] & 2.08 [1.2,4.6] & 240 & 946 \\
4 & 86 [52,125] & 20 [-9,243] & 1.23 [0.2,3.8] & 2.42 [1.4,5.6] & 257 & 1014 \\
5 & 91 [13,143] & 43 [-61,594] & 2.60 [0.3,5.5] & 2.88 [1.6,5.2] & 144 & 564 \\
\midrule
6 & 91 [63,159] & 10 [-3,227] & 0.95 [0.1,6.9] & 2.91 [1.7,6.6] & 479 & 1892 \\
7 & 6 [1,26] & 2 [-1,172] & 0.19 [0.1,2.6] & 3.97 [2.3,10.7] & 848 & 3200 \\
8 & 86 [19,161] & 20 [-22,310] & 1.07 [0.1,5.3] & 1.68 [0.9,4.2] & 326 & 1290 \\
\bottomrule
\end{tabular}
\caption{Pressure $p$, flow rate $q$, cross-sectional area $A$, nodal distance $h$, number of nodes, and number of edges, in the graphs used in the results of \Cref{subsec:convergence} (rows 1-5) and \Cref{subsec:comparison} (rows 6-8). Values of $p$, $q$, $A$, and $h$ are averaged over branch nodes. Minimum and maximum values are reported in brackets.}
\label{table:statistics}
\end{table}

\section{Results}
\label{sec:results}

In this section, we present the results obtained on the datasets described in \Cref{sec:dataset}. We trained the networks in a distributed fashion over 16 36-core dual-socket Intel(R) Ice Lake Xeon(R) nodes. For inference, we use an Apple M1 Max processor.

\subsection{Convergence study with respect to dataset size and sensitivity analysis}
\label{subsec:convergence}

\begin{figure}
\centering
\includegraphics[scale= 0.59]{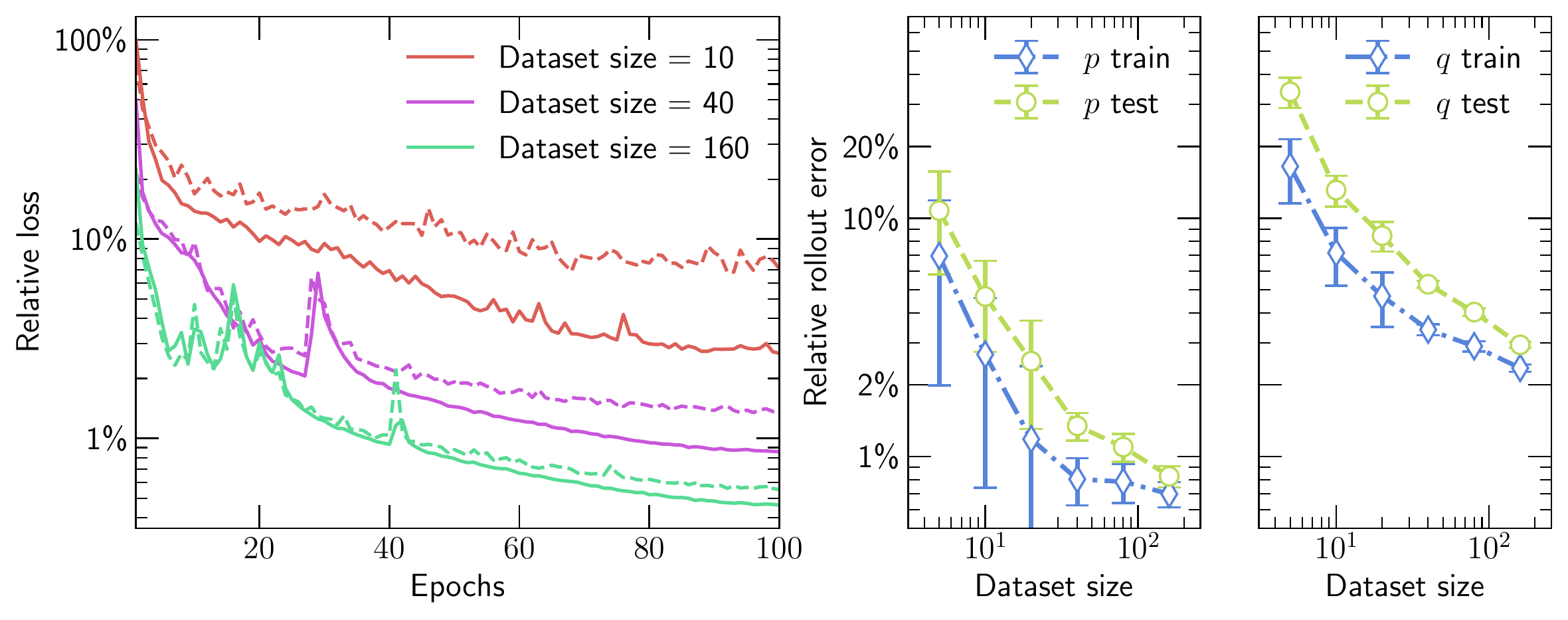}
\caption{MeshGraphNet performance when trained over datasets with variable numbers of unique (i.e., not considering augmented data) trajectories. Left: decay of the loss function over 100 epochs (solid lines: train loss, dashed lines: test loss). The displayed loss is relative to the largest value achieved by the network trained on 10 trajectories. Right: convergence of relative rollout errors in pressure $p$ and flow rate $q$. Vertical bars represent 95\% confidence intervals based on 10 independent training runs.}
\label{fig:convergence}
\end{figure}

We first assessed the performance of the GNN as a function of the dataset size. We used the five healthy models shown in \Cref{fig:dataset} (Models 1-5) and extracted the solution from the last cardiac cycle (32 per geometry) to train MeshGraphNet. Considering data augmentation, this resulted in 640 resolved flows in the dataset. We performed 10-fold cross-validation by considering a 90-10\% split between train and test sets and avoiding data leaks---that is, the augmented data for each simulation belonged to the same (train or test) set as the original simulation.

\Cref{fig:convergence} shows the performance of the network when trained over 10, 20, 40, 80, and 160 trajectories. On the left, we report the train and test loss value decay over the training epochs. As the size of the dataset increases, the gap between train and test validation loss decreases, as does the value of the final loss. On the right, we numerically demonstrate the expected convergence in the pressure and flow rate errors (computed as in \Cref{eq:err_p} and \Cref{eq:err_q}) with respect to the dataset size. Our results also show that, as we consider more trajectories to train the GNN, the generalization gap between train and test rollout errors decreases and is less variable. The variability in performance is represented in \Cref{fig:convergence} (right) by 95\% confidence intervals computed over the 10 networks trained during 10-fold cross-validation.

\begin{figure}
\centering
\includegraphics[scale= 0.59]{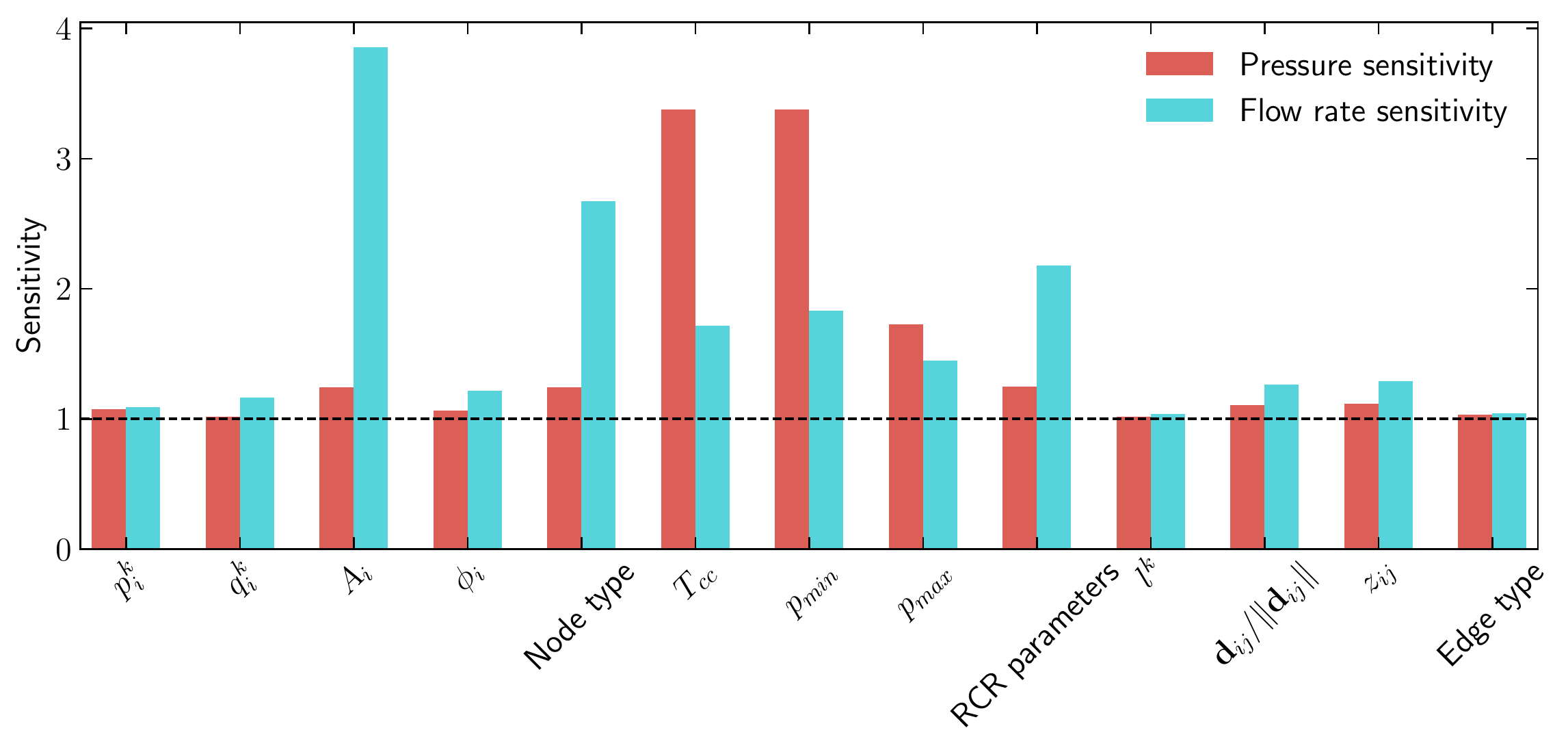}
\caption{Sensitivity analysis of MeshGraphNet with respect to node and edge features. Sensitivity is measured as the ratio between the perturbed and baseline errors, where we obtain the perturbed configuration by adding random Gaussian noise to the features on the x-axis during rollout. Results are averaged over ten independently trained GNNs. The node and edge features are: nodal pressure at time $t^k$ ($p_i^k$), nodal flow rate at time $t^k$ ($q_i^k$), cross-sectional area $A_i$, centerline tangent $\boldsymbol \phi_i$, node type, cardiac cycle period $T_{cc}$, minimum pressure $p_\text{min}$, maximum pressure $p_\text{max}$, RCR parameters, loading variable $l^k$, relative node position $\mathbf{d}_{ij}/\Vert \mathbf{d}_{ij} \Vert$, shortest path length between nodes $n_j$ and $n_i$, and edge type.}
\label{fig:sensitivity}
\end{figure}

We analyzed the sensitivity of MeshGraphNet with respect to each of the node and edge features as follows. We selected one trajectory at random in the dataset and assessed the baseline performance of each of the 10 GNNs trained over the whole dataset in terms of pressure and flow rate rollout errors. Then, we performed many rollouts using the same GNNs while applying random Gaussian noise $N(0, 0.05)$ to only one feature per rollout in each node and edge in each timestep (in the case of multidimensional node features such as tangent and node type, we divide the standard deviation by the size of the vector). We recall that node and edge features are standard normalized, which motivates the use of the same noise distribution for all features. We define the sensitivity factor with respect to a particular feature as the ratio between the error obtained in the perturbed configuration and the baseline error. Therefore, sensitivity close to one indicates robustness with respect to noise. Features associated with higher values of sensitivity are more important to the result accuracy than those associated with lower values. Our findings are summarized in \Cref{fig:sensitivity}. 

The accuracy in the global approximation of pressure and flow rate did not depend dramatically on the previous timestep. As a matter of fact, we train our GNNs specifically to be robust to noise in pressure and flow rate, as discussed in \Cref{subsec:training}. The networks were most sensitive to variations in the cross-sectional area (which is consistent with our understanding of fluid dynamics) and patient-specific data such as boundary conditions parameters and reference values for pressure ($p_\text{min}$ and $p_\text{max}$). Unexpectedly, the cardiac cycle period $T_\text{cc}$ also played an important role; we believe that this is due to the low number of distinct $T_\text{cc}$ considered here (five, i.e., one per geometry) and that increasing the variability of that parameter in the dataset will also reduce its effect. Geometrical features such as tangent or node distance did not significantly affect the accuracy of the results. This might be partially due to the fact that those features are not independent of each other---for instance, the tangent value can be estimated starting from the node positions and vice versa. Finally, we observe that the loading variable $l^k$, which we introduced to distinguish between the loading phase and cardiac cycle simulation, did not play an important role in the rollout accuracy.

\subsection{Comparison against physics-based one-dimensional models}
\label{subsec:comparison}

To assess the performance of our method against a physics-based ROM, we considered Models 6, 7, and 8 (see \Cref{fig:dataset}), which present characteristics that are typically difficult to handle using one-dimensional models. For each geometry, we simulated two cardiac cycles using 50 random configurations of the boundary conditions associated with that cardiovascular model in the VMR.

We also performed one-dimensional simulations using the same boundary conditions used in the three-dimensional ones. We set the material properties of the arterial wall to obtain close-to-zero variations in the lumen area, following the approach adopted in \cite{pfaller2022}. We refer to \Cref{sec:oned} for more details about our implementation of one-dimensional models. We initialized the ROM with the pressure and flow rate approximations by the three-dimensional model evaluated at the one-dimensional nodes and ran it for two cardiac cycles. In our experience, this configuration leads to better agreement with the three-dimensional data (instead of initializing the ROM using constant values for pressure and flow rate or performing the simulation for more cardiac cycles).

Similarly to \Cref{subsec:convergence}, we used 4-fold data augmentation for training and, therefore, each anatomical model was associated with a training dataset of 200 trajectories. We considered two approaches for training. In the first approach, we considered a global dataset with trajectories computed on all three cardiovascular models. We denoted the networks trained this way GNN-A. The second strategy consists of training three distinct GNNs over datasets comprising trajectories from the same anatomical models. In other words, we trained one GNN on all trajectories associated with Model 6, one on all trajectories associated with Model 7, and one on all trajectories associated with Model 8. As these networks are trained on a single geometry, we denote them GNN-B$g$, where $g=6$, $g=7$ or $g=8$ is the patient identifier.  \Cref{table:statistics}, which reports statistics on the models considered in this section, motivates the choice of training geometry-specific networks. Indeed, different vascular regions are characterized by different ranges for key quantities such as pressure, flow rate, and cross-sectional area. Since the GNN operates on data normalized using statistics computed over all graphs in the dataset, we aim to investigate whether normalizing the data with values adapted to each cardiovascular region presents advantages in terms of accuracy. We perform 5-fold cross-validation for GNN-A and GNN-B$g$. Due to the different dataset sizes, we train GNN-A networks for 100 epochs and GNN-B$g$ ones for 500.

\begin{figure}
\centering
\includegraphics[scale=0.59]{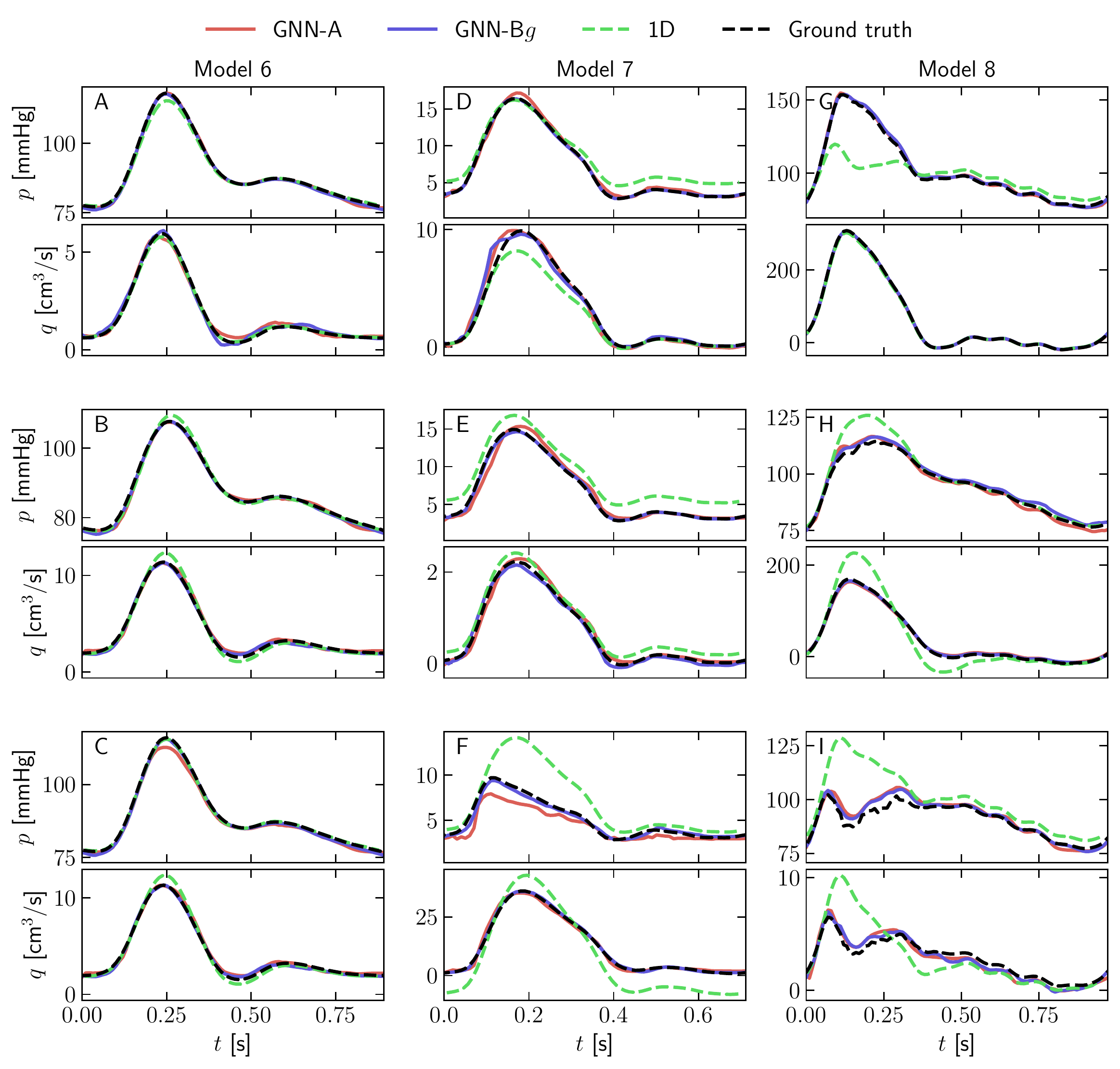}
\caption{Pressure and flow rate approximated by GNNs and one-dimensional models at the locations shown in \Cref{fig:dataset}. Left, middle, and right columns correspond to Models 6, 7, and 8, respectively. All results are obtained with $t = 10^{-2}$ s. In the legend, GNN-A refers to a network trained on all three geometries at the same time, GNN-B$g$ is a GNN trained on simulation trajectories for a single patient ($g = 6$, $g = 7$, or $g = 8$), and 1D refers to the physics-driven one-dimensional model (see \Cref{sec:oned}). Ground truth values are reported in black dashed lines.}
\label{fig:1d-comparison-qual}
\end{figure}

\Cref{fig:1d-comparison-qual} shows the performance of GNN-A, GNN-B6 (left column), GNN-B7 (middle column), and GNN-B8 (right column), over trajectories contained in the test sets. We evaluate the error in pressure at 20 graph nodes sampled at random in the models' branches and display the pressure and flow rate evolution at the locations where the error in pressure achieved by GNN-A is minimum (top row), median (middle row), and maximum (bottom row). The GNNs are remarkably accurate in all considered locations. \Cref{fig:1d-comparison-qual} also shows the performance of one-dimensional models ran with $\Delta t = 10^{-2}$ s (same time step as GNN-A and GNN-B$g$). The data- and physics-driven models both performed well on the aortofemoral model (left column), but the GNNs outperformed the one-dimensional ROM on the pulmonary and aorta coarctation models (middle and right columns).

% \begin{table}[t]
% \centering
% \begin{tabular} {c c c c c c c c c c c}
% \toprule
% & & \multicolumn{3}{c}{Model 6} & \multicolumn{3}{c}{Model 7} & \multicolumn{3}{c}{Model 8}\\
% \cmidrule(lr){3-5}\cmidrule(lr){6-8} \cmidrule(lr){9-11}
% & $\Delta t$ & $e_p$ & $e_q$ & r.t. & $e_p$ & $e_q$ & r.t. & $e_p$ & $e_q$ & r.t.\\
% \midrule
% GNN-A & $0.02$ & 2.5\% & 4.1\% & 1.6 & 3.1\% & 5.1\% & 2.0 & 2.0\% & 3.2\% & 1.5 \\
% GNN-A & $0.01$ & 1.0\% & 2.8\% & 3.1 & 1.4\% & 2.2\% & 4.0 & 1.4\% & 2.7\% & 2.9 \\
% GNN-B$g$ & $0.01$ & 0.7\% & 2.7\% & 3.1 & 1.8\% & 2.2\% & 4.1 & 1.0\% & 2.6\% & 2.8 \\
% \midrule
% 1D & $0.02$ & 3.3\% & 3.0\% & 2.3 & 20.5\% & 17.1\% & 42.4 & 8.8\% & 18.9\% & 0.6 \\
% 1D & $0.01$ & 1.9\% & 3.0\% & 2.7 & 20.5\% & 17.2\% & 60.6 & 8.8\% & 19.6\% & 0.7 \\
% 1D & $0.001$ & 1.6\% & 3.0\% & 15.2 & 20.7\% & 17.4\% & 477.2 & 9.8\% & 20.2\% & 3.4 \\
% \bottomrule
% \end{tabular}
% \caption{Average errors in pressure $e_p$, average errors in flow rate $e_q$, and run time in seconds for different configurations of GNN and one-dimensional models.}
% \label{table:1d-comparison}
% \end{table}

% \label{sec:results}
\begin{figure}
\centering
\includegraphics[width = 1\textwidth]{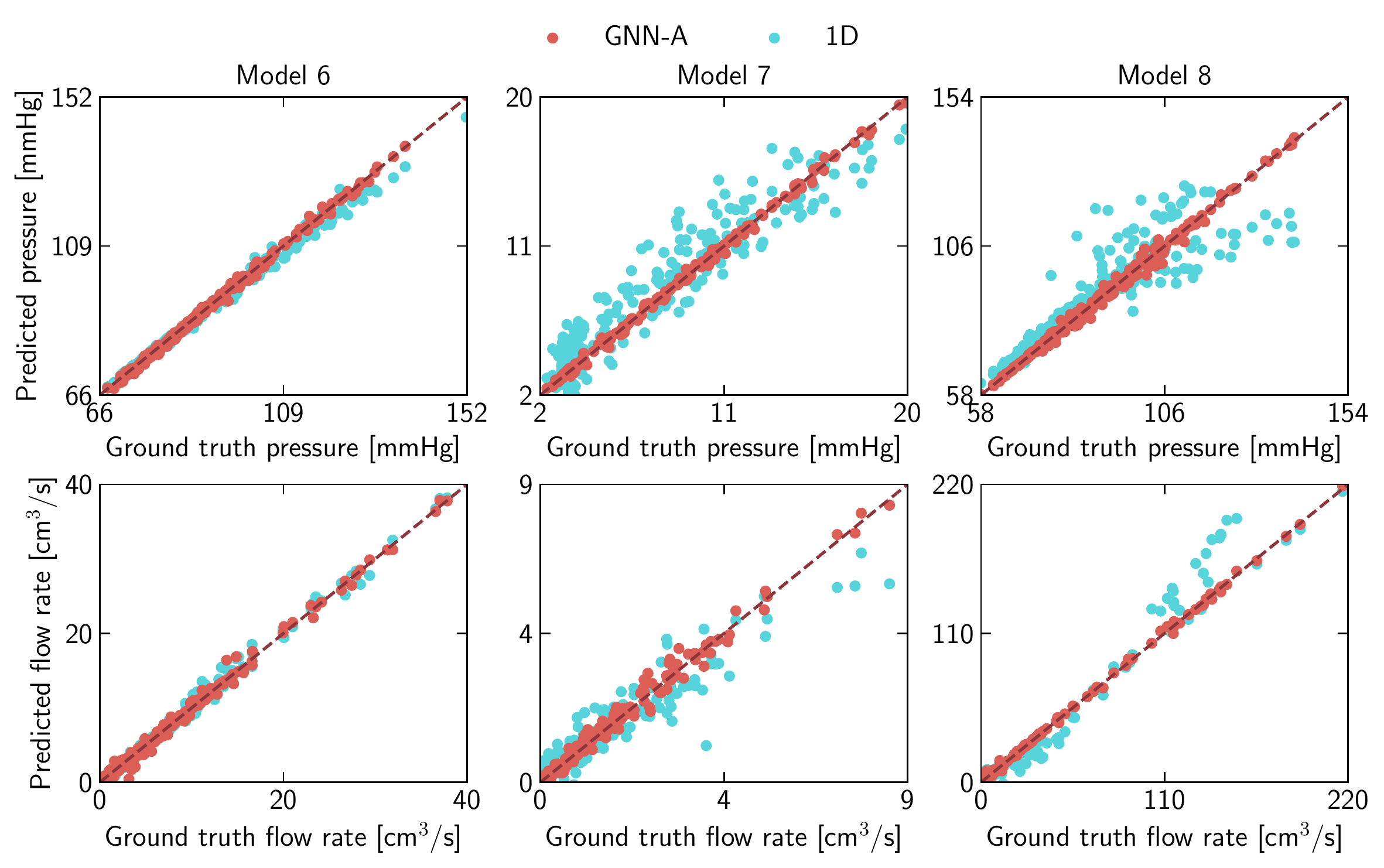}
\caption{Pressure and flow rate approximations by GNN-A and one-dimensional models with $\Delta t = 10^{-2}$ (y-axis) vs ground truth (x-axis). Left, middle, and right column refer to Model 6, 7 and 8, respectively. Points are randomly sampled in space and time over all trajectories in the dataset.}
\label{fig:scatter}
\end{figure}

% \caption{Average errors in pressure and flow rate $e_p$ and $e_q$ for different configurations of GNN and one-dimensional models.}

\Cref{fig:scatter} compares the accuracy of the GNN-A and one-dimensional models over the whole dataset; since GNN-A and GNN-B$g$ performed comparatively well, we here focus on the more general GNN-A for simplicity. In all cases $\Delta t = 10^{-2}$ s. We plot the pressure (top row) and flow rate (bottom row) approximations by the GNNs and one-dimensional models against the ground truth at locations randomly sampled in space and in time in all the models in the dataset. Points positioned on the bisector indicate good agreement between ROM approximation and ground truth. Once again, these results demonstrate that the GNN achieves a much higher accuracy than the physics-based one-dimensional models.

\begin{figure}
\centering
\includegraphics[scale=0.59]{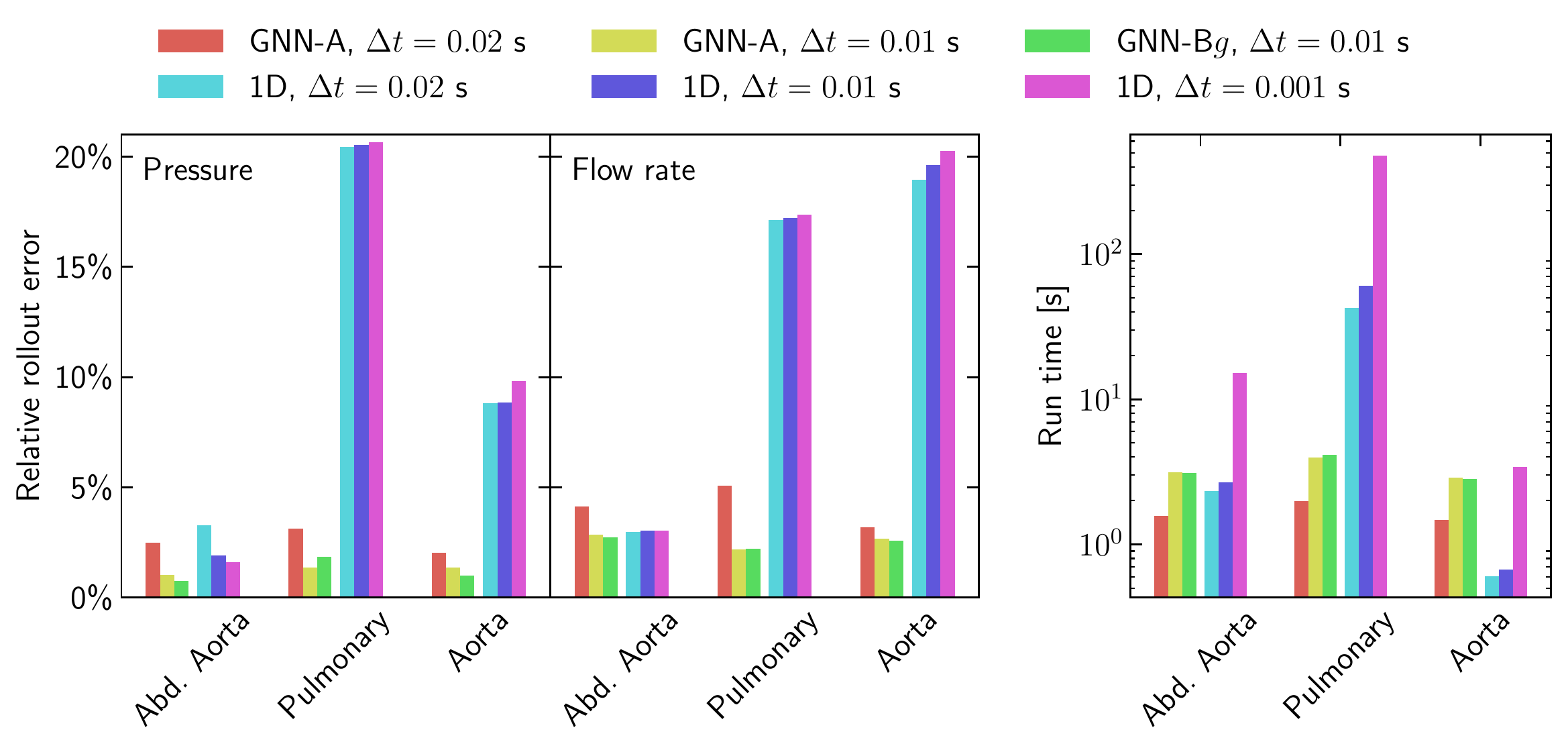}
\caption{Average errors in pressure and flow rate (left panel) and average run time in seconds (right panel) for different configurations of GNNs and one-dimensional models.}
\label{fig:performance}
\end{figure}

In \Cref{fig:performance}, we report the average performance of the GNNs and one-dimensional models over all 150 trajectories. Owing to 5-fold cross-validation and the fact that each trajectory appears in the test set exactly once, the GNN results are averaged over all trajectories in the dataset. We observe that the accuracy of GNN-A and GNN-B$g$ are comparable, which suggests that a single GNN can generalize to multiple geometries. We also note that using $\Delta t = 2 \cdot 10^{-2}$~s instead of $\Delta t = 10^{-1}$~s when training GNN-A leads to lower accuracy. However, we highlight that different time-step sizes might require a different set of hyperparameters---most notably, the standard deviation of the noise we use to make the network robust during rollout. In our case, the hyperparameter optimization discussed in \Cref{subsec:hpo} is performed considering $\Delta t = 10^{-2}$ s. 

The relative errors for pressure and flow rate produced by the GNNs in each anatomical model are considerably lower than those produced by the one-dimensional ROMs. Moreover, we do not observe a strong effect of $\Delta t$ on the global accuracy of one-dimensional models. This indicates that the poor performance achieved on Models 7 and 8 was due to limitations intrinsic to the methods. 

The GNNs showed similar efficiency than the one-dimensional models in terms of run time using the same time step size $\Delta t$ in the Model 6 and lower efficiency in Model 8. In the case of the more complex pulmonary model (Model 7), the GNNs are orders of magnitude more efficient than the one-dimensional models: for example, for $\Delta t = 0.01$ s the one-dimensional models took 60.6 seconds to complete a single cardiac cycle, whereas GNN-A and GNN-B$g$ took around 4.0 seconds. 

We observe that, when using a larger $\Delta t$ to train GNN-A, the run time decreased accordingly: the run times for Models 6, 7, and 8 scales from 3.1, 4.0, and 2.9 (approximately the same for GNN-A and GNN-B$g$) to 1.6, 2.0, and 1.5, respectively. The run times in the one-dimensional models do not scale linearly due to the larger number of Newton iterations necessary to reach convergence in the solution of the nonlinear system with larger $\Delta t$. 

\begin{figure}
\centering
\includegraphics[scale=0.59]{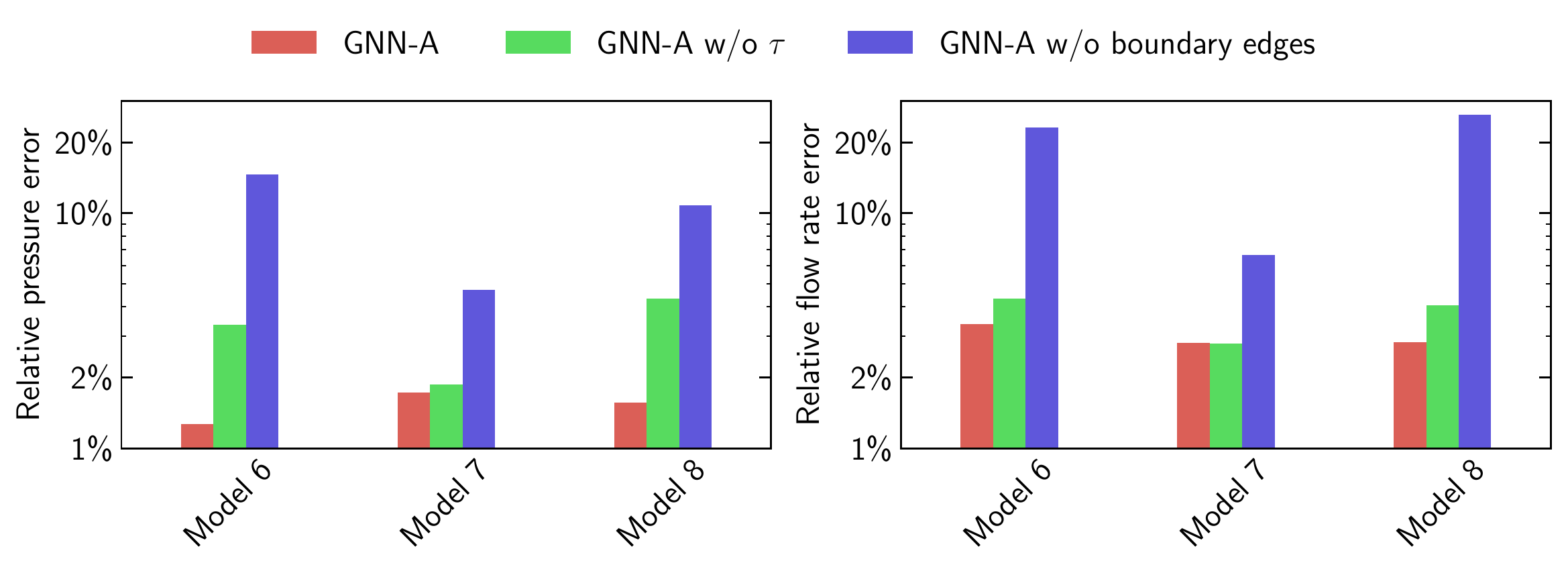}
\caption{Ablation study. We show the decline in performance in pressure (left) and flow rate (right) when we exclude the features defined in \Cref{eq:tau} from the graph (green bars), or when we do not add boundary edges (blue bars), compared to the baseline GNN-A (red bars).}
\label{fig:ablation}
\end{figure}

Finally, we performed an ablation study to evaluate the effects of different GNN components by excluding them altogether. Our goal was to determine whether the main modifications to MeshGraphNet proposed in this paper improve the accuracy of the original algorithm. These modifications are: (i) all graph features listed in \Cref{eq:node_features} and \Cref{eq:edge_features}, except for $p_i^k$, $q_i^k$, $\boldsymbol{\alpha}_i$, $\mathbf{d}^T_{ij}/\Vert \mathbf{d}^T_{ij} \Vert$, and $z_{ij}$, and (ii) the boundary edges discussed in \Cref{subsec:bcs}. For the sake of clarity, we define the set
\begin{equation}
    \tau = \{A_i, \boldsymbol{\phi}_i, T_\text{cc}, p_\text{min}, p_\text{max}, R_{i,p}, C_i, R_{i,d}, l^k, \boldsymbol{\beta}_{ij},\text{ for all }i,j,k\},
    \label{eq:tau}
\end{equation}
which includes all node and edge graph features we propose to incorporate in MeshGraphNet, for all nodes, edges, and timesteps.
We trained one GNN-A by excluding $\tau$ and another one by excluding boundary edges using the same hyperparameters discussed in \Cref{subsec:hpo}, performed 5-fold cross-validation, and compared their performance against a baseline GNN-A. \Cref{fig:ablation} shows the results averaged over all networks and divided into Models 6, 7, and 8. We observed the biggest performance decline when excluding the boundary edges. This suggests that, in our application, MeshGraphNet cannot be used as is without including those edges to allow information to flow more quickly in the graph. Excluding all the graph features we introduced in this paper also resulted in a noticeable performance drop, particularly in the diseased models (Models 6 and 8). These results demonstrate that our modifications to MeshGraphNet lead to noticeable improvements in the ROM.

\section{Conclusions}
\label{sec:conclusions}
We presented a reduced-order model to simulate blood dynamics in one-dimensional approximations of patient-specific vasculatures. Our architecture is a modified version of MeshGraphNet adapted to suit cardiovascular simulations. We demonstrated the generalizability of the network on a variety of different geometries and topologies. We showed the convergence of rollout error on train and test datasets as the number of trajectories used for training increased. We performed a sensitivity analysis to determine which node and edge features are more important to correctly approximating pressure and flow rate. In our experiments, the most influential features were nodal cross-sectional area and patient-specific quantities such as parameters governing the boundary conditions. In \Cref{subsec:comparison}, we carried out a direct comparison with physics-driven one-dimensional models, showing superior performance of the graph neural network, in particular when handling complex geometries such as those with many junctions (Model 7) or stenosis (Model 8). Specifically, our networks consistently achieved errors below 2\% and 3\% in pressure and flow rate, respectively. We also considered different approaches to train our algorithm: training networks specific to different cardiovascular regions instead of a single network able to handle different geometries. Our results show that low errors can be obtained by following the latter strategy, which indicates that, with sufficient training data, our algorithm will adequately mimic three-dimensional simulations in all regions of the cardiovascular tree. Motivated by our sensitivity study, we performed an ablation study to determine which of our contributions played a more critical role in the accuracy of the network when excluding them altogether. Our results indicate that introducing boundary edges is essential to ensure meaningful results and that using patient-specific graph features also leads to performance improvements with respect to the original MeshGraphNet architecture.

Future work will focus on further quantifying and improving the graph neural network's ability to generalize to unseen geometries during training. We will pursue methods that enable the graph neural network to robustly achieve the accuracies and efficiencies demonstrated in this work using smaller training datasets. This could be done by modifying the structure of the graph neural network (e.g., by incorporating notions of physics like conservation of mass) or the composition of the training dataset to include more patient-specific geometries with fewer trajectories each.

Another open direction of research (motivated by our sensitivity analysis in \Cref{subsec:convergence}) consists of investigating whether introducing new features or removing some of the existing ones will improve the performance of the method. Removing the parameters associated with the boundary conditions---while keeping the patient-specific data that is normally used to determine them---would be a significant advantage over existing approaches. Indeed, boundary condition tuning is a critical step in current physics-based simulations and is typically performed by varying the boundary condition parameters of surrogate models (for example, zero- or one-dimensional reduced-order models) in Bayesian optimization frameworks that usually require many model evaluations to converge. These optimization procedures are based on objective functions that measure how closely the surrogate model reproduces key physiological quantities such as systolic and diastolic pressures. Incorporating these physiological measures into the neural networks would allow us to bypass the boundary condition tuning stage and reduce the number of steps between medical image acquisition and simulation result assessment. The results presented in \Cref{sec:noRCR} give us confidence that further tuning of the model and improvements in the training dataset will allow us to obtain accurate approximation results without providing boundary condition parameters to the networks.

\section*{Acknowledgments}

This work was supported by NIH Grants R01LM013120, R01EB029362, and K99HL161313.  Additional funding was provided by the Stanford Graduate Fellowship and an NSF GRFP. This publication was additionally supported by the Stanford Maternal and Child Health Research Institute. The authors gratefully acknowledge the San Diego Super Computing Center (SDSC) and Intel for providing the computational resources to run the three-dimensional simulations and to train the GNNs presented in this paper. The authors also thank Dr.\ Tailin Wu for the insightful discussions and support on GNNs implementation and calibration.

\bibliographystyle{unsrtnat}
\bibliography{references}

\appendix
\section{Details on one-dimensional models}
\label{sec:oned}
One-dimensional models approximate pressure, flow rate, and wall displacement along the centerline of compliant blood vessels. Given a curve in three-dimensional space parametrized by axial variable $z$, we define  $p=p(z,t)$, $q=q(z,t)$ and $A=A(z,t)$ are pressure and flow rate of blood and the area of the vessel lumen at $z$, respectively. Under the assumption of Poiseuille flow, the blood dynamics equations read
\begin{equation}
\begin{aligned}
    \dfrac{\partial A}{\partial t} + \dfrac{\partial q}{\partial z} &= 0,\\
    \dfrac{\partial q}{\partial t} + \dfrac{\partial }{\partial z}\left (\dfrac{4}{3} \dfrac{q^2}{A} \right ) &= -8\pi \nu \dfrac{q}{A} + \nu \dfrac{\partial^2 q}{\partial z^2} - \dfrac{A}{\rho} \dfrac{\partial p}{\partial z},
    \label{eq:oned}
\end{aligned}
\end{equation}
where $\nu$ is the kinematic viscosity of blood, which is typically set to $\nu = 3.77 \cdot 10^{-2}\,\text{s}\,\text{cm}^{-2}$. To accurately predict the vessel response to blood flow, it is necessary to supplement \Cref{eq:oned} with a constitutive model. The Olufsen model
\[
    p(z,t) = p_0(z) + \dfrac{4}{3}\left (k_1 e^{k_2 r_0(z)} + k_3\right )\left (1 - \sqrt{\dfrac{A_0(z)}{A(z,t)}}\right )
\]
is an example of commonly used constitutive law based on a reduced form of linear elasticity \cite{olufsen1999structured}. Here, $p_0(z)$, $r_0(z)$, $A_0(z)$ are reference values for pressure, radius, and lumen area, and $k_1$, $k_2$, and $k_3$ are empirical constants. Since the three-dimensional simulations we use as benchmark are obtained under rigid-wall assumptions, in the tests presented in \Cref{subsec:comparison} we set $k_1 = 0$ and $k_3$ to an arbitrarily large value, to obtain the same kind of wall response in the one-dimensional models. This approach was also followed in \cite{pfaller2022}, where one-dimensional models were compared with three-dimensional simulations on 72 diverse cardiovascular geometries.

While \Cref{eq:oned} models behavior in branches, we need to add conditions to model the dynamics in junctions. Typical conditions include flow rate conservation---i.e., the sum of incoming inflow must equal the sum of outgoing outflow---and zero pressure drop between junction inlets and outlets.

Our results are obtained using an open-source C++ code publicly available at \url{https://github.com/SimVascular/svOneDSolver}. 

\section{Cardiovascular models reference identifiers}

The cardiovascular models used in this study are associated with the following identifiers in the VMR: $1 = 0090\_0001$, $2 = 0091\_0001$, $3 = 0093\_0001$, $4 = 0094\_0001$, $5 = 0095\_0001$, $6 = 0140\_2001$, $7 = 0080\_0001$, $8 = 0104\_0001$.

\section{Removal of boundary conditions parameters from graph features}
\label{sec:noRCR}

\begin{figure}
\centering
\includegraphics[scale=0.59]{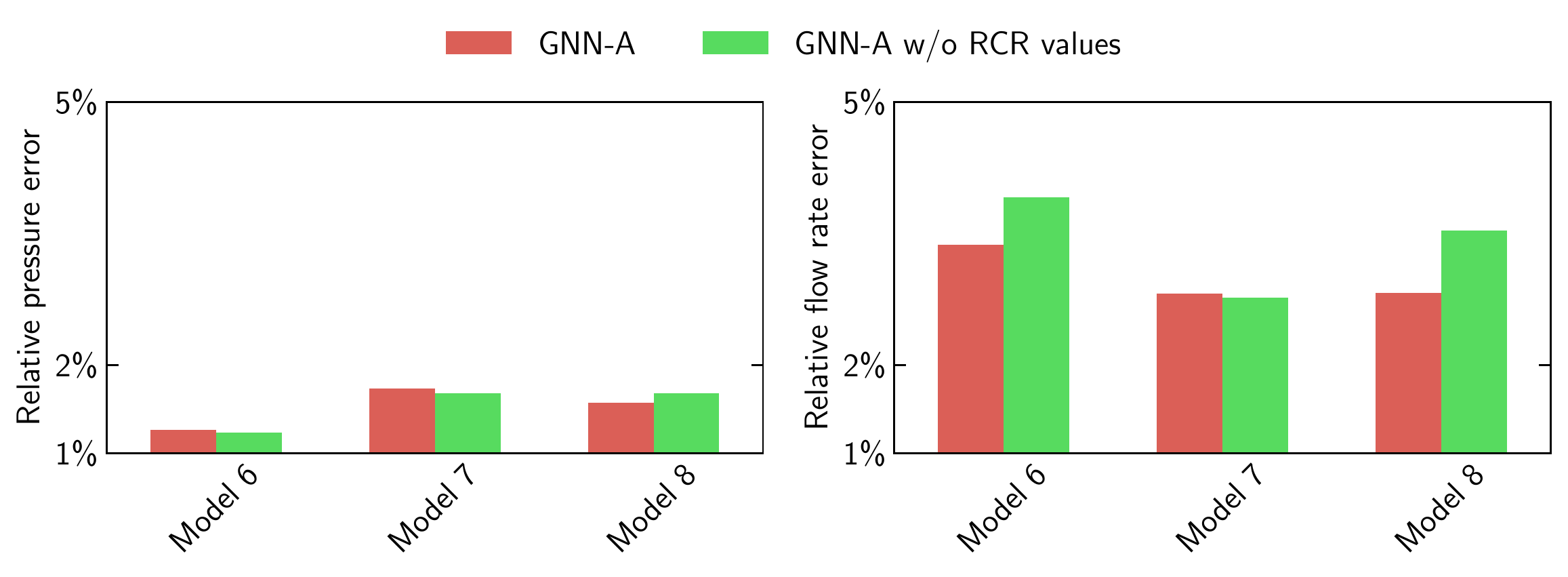}
\caption{Red bars: baseline GNN-A trained on Model 6, Model 7, and Model 8. Green bars: modified GNN-A where RCR values are not included in the set of node features at the boundary nodes.}
\label{fig:ablation_noRCR}
\end{figure}

Similar to the ablation study performed in \Cref{subsec:comparison}, we aimed to investigate the effects of removing a feature from the graph and evaluating the effects on the GNN performance. In particular, we focused on RCR values for boundary conditions. This choice is motivated by the fact that, in the standard simulation pipeline, RCR parameters are estimated starting from physiological measurements in an optimization process that requires numerous model evaluations. Due to the flexibility provided by data-driven approaches that allows the easy incorporation of physiological features during inference (in our case, for example, we include  $p_\text{min}$ and $p_\text{max}$ among the graph node features, which are often used to estimate boundary values parameters), we foresee that, in the future, it will be possible to omit the boundary condition tuning step in the simulation pipeline. To motivate this hypothesis, we trained a GNN-A by removing RCR values from the node graph features. We report the performance obtained on the same three models considered in \Cref{subsec:comparison} in \Cref{fig:ablation_noRCR}. Similar to the results presented in \Cref{fig:ablation}, these results are averages obtained during 5-fold cross-validation. Although the addition of RCR values to the graph led to increased accuracy in some cases (particularly in the flow rate approximation), the difference with respect to the baseline GNN-A was negligible, which supports our initial hypothesis. Future work will focus on corroborating our hypothesis further and on analyzing the dependencies among graph features to decrease their number while preserving accuracy.

\end{document}